\documentclass[twoside,11pt]{article}

\usepackage[abbrvbib, preprint]{jmlr2e}

\usepackage{jmlr2e}
\usepackage{algorithm}  
\usepackage{algorithmic}
\usepackage{bbm}
\usepackage{subfigure}
\usepackage{multirow} 
\usepackage{amsmath}
\usepackage{mathtools}

\jmlrheading{1}{2021}{1-48}{4/00}{10/00}{meila00a}{Qiuqiang Lin}

\ShortHeadings{Random Intersection Chains}{Lin}
\firstpageno{1}

\begin{document}

\title{Random Intersection Chains}

\author{\name Qiuqiang Lin \email 11735034@zju.edu.cn \\
       \addr School of Mathematical Sciences\\
       Zhejiang University\\
       Hangzhou 310027, China
       \AND
       \name Chuanhou Gao \email gaochou@zju.edu.cn \\
       \addr School of Mathematical Sciences\\
       Zhejiang University\\
       Hangzhou 310027, China}

\editor{}

\maketitle

\begin{abstract}
  Interactions between several features sometimes play an important role in prediction tasks.
  But taking all the interactions into consideration will lead to an extremely heavy computational burden.
  For categorical features, the situation is more complicated since the input will be extremely high-dimensional and sparse if one-hot encoding is applied.
  Inspired by association rule mining, we propose a method that selects interactions of categorical features, called Random Intersection Chains.
  It uses random intersections to detect frequent patterns, then selects the most meaningful ones among them.
  At first a number of chains are generated, in which each node is the intersection of the previous node and a random chosen observation.
  The frequency of patterns in the tail nodes is estimated by maximum likelihood estimation, then the patterns with largest estimated frequency are selected. 
  After that, their confidence is calculated by Bayes’ theorem.
  The most confident patterns are finally returned by Random Intersection Chains.
  We show that if the number and length of chains are appropriately chosen, the patterns in the tail nodes are indeed the most frequent ones in the data set.
  We analyze the computation complexity of the proposed algorithm and prove the convergence of the estimators.
  The results of a series of experiments verify the efficiency and effectiveness of the algorithm.
\end{abstract}


\section{Introduction}
\label{sec:introduction}
For practical application of machine learning algorithms, usually not only the original features, 
but also their interactions play an important role.
However, taking all the interactions into consideration will lead to an extremely high-dimensional input.
For example, even if only the product of two numerical features is considered,
there will be $O(p^2)$ such interactions for $p$ main effects.
As the size of data is growing rapidly nowadays, adding all pairwise interaction into the input may be computationally infeasible, let alone higher-order interactions.
What's more, adding all the interactions without selection is possibly harmful to the prediction model since too much redundant information is brought in.

It is a well-established practice among statisticians fitting models on interactions as well as original features.
For example, 
\cite{bien13} add a set of convex constraints to the lasso that honour the hierarchy restriction;
\cite{hao14} tackle the difficulty by forward-selection-based procedures;
\cite{hao18} consider two-stage LASSO and a new regularization method named \textit{RAMP} to compute a hierarchy-preserving regularization solution path efficiently;
\cite{agrawal19} propose to speed up inference in Bayesian linear regression with pairwise interactions by using a Gaussian process and a kernel interaction trick.
However, these methods are based on the hierarchy assumption.
That is, an interaction will be useful only if its lower-order components are also useful.
So the theoretical analysis lose efficacy and their practical performance may be unsatisfactory for the case where the assumption does not hold. 

There is also some work free of the hierarchy assumption. 
\cite{Thanei18} propose the \textit{xyz} algorithm, 
where the underlying idea is to transform interaction search into a closest pair problem which can be solved efficiently in subquadratic time. 
Instead of the hierarchy principle, \cite{yu19} come up with the reluctant principle, which says that one should prefer main effects over interactions given similar prediction performance.

The above-mentioned work mainly aims to select pairwise interactions.
A drawback of these methods is that potentially informative higher-order interactions are overlooked.
\textit{Random intersection trees} \citep{shah14} gets over this difficulty by starting with a maximal interaction that includes all variables, and then gradually removing variables if they fail to appear in randomly chosen observations of a class of interest. 
Another approach is \textit{Backtracking} \citep{shah16}.
It can be incorporated into many existing high-dimensional methods based on penalty functions, and works by building increasing sets of candidate interactions iteratively.

An alternative approach for higher-order interaction selection is extracting interactions from rules.
Decision trees, such as ID3 \citep{Quinlan86}, C4.5 \citep{Quinlan93} and CART \citep{Breiman84}, are widely-used models to generate comprehensive rules.
They work by partitioning the input space according to whether the features satisfy some conditions, 
and then assigning a constant to each region. 
After the tree is built, each path connecting the root node and a leaf node can be transformed into a decision rule by combining the split decisions. 
The predictions of the leaf nodes are discarded and only the splits are used in the decision rules.
\textit{RuleFit} \citep{Friedman08} uses these decision rules as binary features, then fits a linear model on them.
According to \cite{qu16}, however, the exploration ability of tree-based models is restricted for high-dimensional categorical features due to the low usage rate of categorical features.


Besides tree-based models, there is another family of algorithms that generate rules based on the data, namely association rule mining.
For a categorical feature $X$, and one of its optional values $x$, we can treat the pattern ``$X=x$'' as an item.
In this way we can transform an observation in the data set to a record of items.
Then association rule mining algorithms can be applied.
Mining association rules between items from a large database is one of the most important and well researched topics of data mining. 
Let $I=\{i_1, i_2, ..., i_m\}$ be a set of items, called itemsets.
Association rule mining aims to extract rules in the form of ``$X\to Y$'', where $X\subset I$, $Y\subset I$, $X\cap Y=\emptyset$. 
Calling $X$ the antecedent and $Y$ the consequent, the rule means $X$ implies $Y$.
The support of an itemset $X$ is the number of records that contain $X$.
For an association rule ``$X\to Y$'', its support is defined as the fraction of records that contain $X\cup Y$ to the total number of records in the database,  
and its confidence is the number of cases in which the rule is correct relative to the number of cases in which it is applicable, or equivalently, support($X\cup Y$)/support($X$).

Association rule mining problem was firstly stated by \cite{agrawal93}, and further studied by many researchers.
Apriori \citep{agrawal94}, FP-growth \citep{han00} and H-mine \citep{pei01} are some well-known algorithms for association rule mining.
If the antecedents is meaningful for the target, it seems reasonable to use them as features for another classification model rather than a classifier themselves, just like how \textit{RuleFit} makes use of decision rules.

Although the common association rule mining algorithms are much faster than a brute force search, many of them are not suitable for ``big data'' since they have to go through the whole database multiple times.
On the contrary, \textit{Random Intersection Trees} gets over this difficulty by regarding the intersection of a set of random samples as a frequent pattern.
To be more specific, it generates $M$ trees of depth $D$, in which each node but the leaf nodes has $B$ child nodes, where $B$ subjects to a pre-specified distribution.
The root node contains the items in a uniformly chosen observation from the database, and each node except the root consists of the intersection of its parent node and a uniformly chosen observation.
The patterns in the leaf nodes are finally used as interactions.
One of the shortcomings of \textit{Random Intersection Trees} is that the selection is relatively crude because the subpatterns of the found frequent patterns are neglected, while they are actually more frequent.
What's more, \textit{Random Intersection Trees} aims to find the patterns that are frequent in a class of interest but infrequent in other classes, which are exactly ``confident rules''.
But the precise quantities of ``frequency'' or ``confidence'' are not provided, which makes it difficult to make a more careful comparison among different patterns.
Another problem is that \textit{Random Intersection Trees} can only deal with binary features.
A categorical feature of high cardinality should be one-hot encoded before applying the algorithm, which may result in a very high-dimensional and sparse input.

Inspired by the idea of \textit{Random Intersection Trees}, 
we suggest a method that can select useful interactions of categorical features for classification tasks, called \textit{Random Intersection Chains}.
The road map of \textit{Random Intersection Chains} is listed below.
\begin{enumerate}
  \item Generate chains for different classes separately by random intersections;
  \item Calculate the frequency of the patterns in the tail nodes as well as their subpatterns by maximum likelihood estimation;
  \item Select the most frequent patterns;
  \item Calculate the confidence of the most frequent patterns by Bayes’ theorem;
  \item Select the most confident patterns.
\end{enumerate}

Our main contributions in this paper can be concluded as follows: 
(1) an interaction selection method for categorical features in classification tasks, named \textit{Random Intersection Chains}, is proposed; 
(2) we show that \textit{Random Intersection Chains} can find all the frequent patterns while leaving out the infrequent ones if parameters are appropriately chosen;
(3) the computational complexity, including space complexity and time complexity, are analyzed;
(4) we prove that the estimated frequency and confidence converge to their true values;
(5) a series of experiments are conducted to verify the effectiveness and efficiency of \textit{Random Intersection Chains}.

The rest of the paper is organized as follows. 
In Section~\ref{sec:preliminaries}, a brief introduction of some related contents is given.
In Section~\ref{sec:algorithm} we introduce \textit{Random Intersection Chains}, our algorithm for interactive feature selection, in detail.
This followed by some analyses of computational complexities in Section~\ref{sec:computation_complexity}.
In Section~\ref{sec:convergence}, we theoretically analyze the the convergence of the estimated frequency and confidence.
In Section~\ref{sec:experiments} we report the results of a series of experiments to verify the effectiveness of the algorithm.
Finally this paper is concluded in Section~\ref{sec:conclusion}.
The proofs of some main results are relegated to the Appendix.

\section{Preliminaries}
\label{sec:preliminaries}
In this paper, we consider the classification task involving high-dimensional categorical predictors.
Usually the categorical features have a number of optional values.
A common approach to deal with such features is one-hot encoding, which transforms a categorical feature to a large number of binary variables.
But this method will make the input extremely high-dimensional and sparse.
To avoid this difficulty, label encoding is adopted in this paper, which maps each category to a specific integer.
For example, there may be three categories, namely ``red'', ``green'' and ``blue'', in a feature representing colors.
Then we replace ``red'', ``green'' and ``blue'' with 1, 2 and 3, respectively.
It's worth noting that these integers can only be used to check whether two values are identical or different,
while their numerical relationships should be ignored.

Suppose $C_1, C_2, ..., C_p$ are $p$ categorical features, and $C$ is the set of classification labels.
The given data set is in the form of $D=\{\boldsymbol{X}, \boldsymbol{y}\}$, 
where $\boldsymbol{X}\in \mathbb{N}^{N\times p}$ contains the records of $N$ observations,
$\boldsymbol{y}\in C^N$ indicates the label of these observations.
The $i$-th row of $\boldsymbol{X}$ and the $i$-th component of $\boldsymbol{y}$ are denoted by $X_i$ and $y_i$, respectively.
Suppose $X_i=[c_1, c_2, ..., c_p]$ is an observation in the data set,
then it can be viewed from two aspects.
First if we treat $c_1, c_2, ..., c_p$ as integers, $\boldsymbol{X}_i$ is naturally a vector of dimension $p$.
Or $c_1, c_2, ..., c_p$ can be seen as items, thus $\boldsymbol{X}_i$ is an record consisting of $p$ items.
Therefore, $\boldsymbol{X}$ can be regarded as a data set for machine learning algorithms, or a database for data mining algorithms.

For a variable $C_j$ and one of its possible values $c_j$, we use ``$C_j=c_j$'' to represent $\mathbbm{1}_{\{C_j=c_j\}}$, a binary feature that indicates whether the value of variable $C_j$ is $c_j$.
``$C_j=c_j$'' also stands for an item that only appears in the records where the value of variable $C_j$ is $c_j$.
Similarly, for $\{j_1, j_2, ..., j_k\}\subseteq \{1, 2, ..., p\}$, suppose $C_{j_1}, C_{j_2}, ..., C_{j_k}$ are $k$ variables and $c_{j_1}, c_{j_2}, ..., c_{j_k}$ are one of their corresponding possible values.
Then a pattern $s$=``$C_{j_1}=c_{j_1}, C_{j_2}=c_{j_2}, ..., C_{j_k}=c_{j_k}$'' can be comprehended as a logical expression, a binary feature $\mathbbm{1}_{\{s\subseteq X\}}$, or an itemset containing $k$ items.
We call such a expression ``$k$-order interaction''.
This definition coincides with the term ``interaction'' used by \cite{shah14}, and will reduce to the latter if $c_j$=1 for all $j\in \{j_1, j_2, ..., j_k\}$.
Also the interaction defined here is a non-additive interaction \citep{Friedman08, Sorokina08, Tsang18}, since it can not be represented by a linear combination of lower-order interactions.
In this paper, we use the terms ``interaction'', ``itemset'' and ``pattern'' interchangeably to describe such expressions.
When there is no ambiguity for classification label $c$, this expression is also referred to ``$s\to c$'' as a ``rule''.

The frequency of an interaction $s$ for class $c$ is defined as the ratio of records containing $s$ with label $c$ to all the records with label $c$, and is denoted by 
\begin{equation}
  p_s^{(c)}=\mathbb{P}_N(s\subseteq X|Y=c)\coloneqq \frac{1}{|I^{(c)}|}\sum_{i\in I^{(c)}}\mathbbm{1}_{\{s\subseteq X_i\}},
\end{equation}
where $I^{(c)}$ is the set of observations in class $c$.
The confidence  of an interaction $s$ for class $c$ is defined as the ratio of records containing $s$ with label $c$ to all the records containing $s$, and is denoted by 
\begin{equation}
  q_s^{(c)}=\mathbb{P}_N(Y=c|s\subseteq X)\coloneqq \frac{1}{|I_s|}\sum_{i\in I^{(c)}}\mathbbm{1}_{\{s\subseteq X_i\}},
\end{equation}
where $I_s$ is the set of observations containing interaction $s$.
An interaction is said to be frequent if its frequency is large, and confident if its confidence is large for a class.

The main goal of this paper is to efficiently detect the interactions that are both frequent and confident for some classes. 
Then these interactions are used as the input for a succeeding classification model.
Since irrelevant predictors in the original input are dropped and useful high-order interactions are explicitly added, interaction detection is likely to be beneficial for the prediction performance.

\section{Random Intersection Chains}
\label{sec:algorithm}
In this section we give a naive version of \textit{Random Intersection Chains} at first, and show that there exist appropriate parameters for it to find all the frequent patterns while the infrequent ones are left out.
Then a modification is provided to prevent the exponential computational burden for selecting the most frequent subpatterns from a given pattern.

\subsection{A Basic Algorithm}
Drawing inspiration from association rule mining and \textit{Random Intersection Trees}, we suggest an algorithm that can efficiently detect useful interactions, named \textit{Random Intersection Chains}.
Like other data mining algorithms, frequent itemsets are discovered at first and then confident rules are generated. 
But instead of scanning the complete database, we adopt random intersections to mine the frequent itemsets, then their frequency and confidence are calculated with the assistant of maximum likelihood estimation or Bayes’ theorem.

The first node of a chain, called the head node, contains the items in a randomly chosen instance.
The other nodes in the chain contain the intersection of its previous node and a new randomly chosen instance.
We repeatedly choose random instances until the length of a chain reaches the pre-defined threshold, 
or the number of the items in the last node (named the tail node) is sufficiently small.
Finally the itemsets in the tail nodes as well as their subsets are regarded as frequent itemsets.
For example, if we want to generate a chain consisting of three nodes, 
and the chosen instances are $[c_1, c_2, c_3]$, $[c_1, c_2', c_3]$, $[c_1, c_2, c_3']$, where $c_1\neq c_1'$, $c_2\neq c_2'$ and $c_3\neq c_3'$, 
then the chain is $[C_1=c_1, C_2=c_2, C_3=c_3]\to [C_1=c_1, C_3=c_3]\to [C_1=c_1]$.
The procedure of generating $M$ such chains of length $D$ can be seen straightly from Algorithm \ref{alg:generate_chain}.

\begin{algorithm}
  \caption{GenerateChain: generate chains by intersection}
  \label{alg:generate_chain}
  \begin{algorithmic}[1]
  \REQUIRE $\{(X_i, y_i)\}_{i\in I^{(c)}}$(observations in class $c$); \\
  D(length of a chain); \\
  M(number of chains);\\
  \ENSURE chains for class c;
  \FOR{$m=1~to~M$}
    \STATE{Draw a random observation $X_{i_1}$ from the observations}
    \STATE{$S_{1,m}^{(c)}\leftarrow X_{i_1}$}
    \FOR{$d=2~to~D$}
      \STATE{Draw a random observation $X_{i_d}$ from the observations}
      \STATE{$S_{d,m}^{(c)}\leftarrow S_{d-1,m}^{(c)}\cap X_{i_d}$}
    \ENDFOR
  \ENDFOR
  \STATE{return $\{\{S_{d,m}^{(c)}\}_{d=1}^D\}_{m=1}^M$}
  \end{algorithmic}
\end{algorithm}

The larger frequency a pattern has, the more likely it is to appear in a uniformly chosen instance.
Therefore, it's reasonable to assume the pattern in tail nodes are more frequent than others.
After detecting the frequent patterns, \cite{shah14} adopt an estimator based on min-wise hashing to obtain their frequency.
Though this estimator enjoys reduced variance compared to that which would be obtained using subsampling, 
it seems somewhat redundant and unnatural because it's independent of \textit{Random Intersection Trees}.
However, \textit{Random Intersection Chains} can estimate frequency by themselves.

For a pattern $s$ with frequency $p_s^{(c)}$, 
denote the number of its appearance in the $m$-th chain for class $c$ by $k_{s,m}^{(c)}$.
The likelihood of observing this chain is
\begin{equation}
  \mathbb{P}(k_{s,m}|p_s)=\begin{cases}
                        p_s^{k_{s,m}}(1-p_s), \hfill \mbox{if~} k_{s,m}<D\\
                        p_s^{k_{s,m}}, \hfill \mbox{if~} k_{s,m}=D
                      \end{cases},
\end{equation}
where we omit the superscript ``$(c)$'' to keep notation uncluttered.
And we have the likelihood of observing $M$ chains as shown in Equation \ref{eq:likelihood},
\begin{equation}
\label{eq:likelihood}
  \begin{aligned}
    \mathbb{P}(\{k_{s,m}\}_{m=1}^M|p_s)&=\prod_{m:k_{s,m}<D}p_s^{k_{s,m}}(1-p_s)\cdot \prod_{m:k_{s,m}=D}p_s^{k_{s,m}}\\
    &=p_s^{K_s}(1-p_s)^{I_s},
  \end{aligned}
\end{equation}
where $K_s=\sum_{m=1}^Mk_{s,m}$, $I_s=\sum_{m=1}^M\mathbbm{1}_{\{k_{s,m}<D\}}$.
Thus the log of likelihood is
\begin{equation}
\label{eq:loglikelihood}
   \log \mathbb{P}(\{k_{s,m}\}_{m=1}^M|p_s)=K_s\log p_s+I_s\log (1-p_s).
\end{equation}
Setting the derivative of Equation \ref{eq:loglikelihood} with respect to $p_s$ equalling to zero and rearranging, we obtain
\begin{equation}
  \hat{p}_s=\frac{K_s}{K_s+I_s}=\frac{\bar{k}_s}{\bar{k}_s+\bar{\chi}_s},
\end{equation}
where $\bar{k}_s=\frac{1}{M}\sum_{m=1}^Mk_{s,m}$, $\bar{\chi}_s=\frac{1}{M}\sum_{m=1}^M\mathbbm{1}_{\{k_{s,m}<D\}}$.

Algorithm \ref{alg:frequency} estimates the frequency of a pattern by maximum likelihood estimation based on \textit{Random Intersection Chains}.
Algorithm \ref{alg:confidence} provides an estimator of confidence by Bayes’ theorem once the frequency is available.

\begin{algorithm}
  \caption{Frequency: estimate the frequency according to chains}
  \label{alg:frequency}
  \begin{algorithmic}[1]
  \REQUIRE $s$(a pattern); \\
  $\{\{S_{d,m}^{(c)}\}_{d=1}^D\}_{m=1}^M$(the chains); \\
  \ENSURE $\hat{p}_s^{(c)}$(the frequency);
  \FOR{$m=1~to~M$}
    \STATE{$k_{s,m}^{(c)}\leftarrow \mathop{\arg\max}_{d}\{s\in S_{d,m}^{(c)}\}$}
    \STATE{$\chi_{s,m}^{(c)}\leftarrow \mathbbm{1}_{\{k_{s,m}^{(c)}<D\}}$}
  \ENDFOR
  \STATE{$\hat{p}_s^{(c)}\leftarrow \frac{\bar{k}_s^{(c)}}{\bar{k}_s^{(c)}+\bar{\chi}_s^{(c)}}=\frac{\sum_{m=1}^M{k_{s,m}^{(c)}}}{\sum_{m=1}^M[{k_{s,m}^{(c)}}+{\chi_{s,m}^{(c)}]}}$}
  \STATE{return $\hat{p}_s^{(c)}$}
  \end{algorithmic}
\end{algorithm}

\begin{algorithm}
  \caption{Confidence: estimate the confidence by Bayes' theorem}
  \label{alg:confidence}
  \begin{algorithmic}[1]
  \REQUIRE $\{\hat{p}_s^{(c)}\}_{c\in C}$(frequency); \\
  $\{p^{(c)}\}_{c\in C}$(prior probabilities); \\
  \ENSURE $\hat{q}^{(c)}$(the confidence);
  \STATE{$\hat{q}_s^{(c)}\leftarrow \frac{\hat{p}_s^{(c)}p^{(c)}}{\sum_{c'\in C}\hat{p}_s^{(c')}p^{(c')}}$}
  \STATE{return $\hat{q}_s^{(c)}$}
  \end{algorithmic}
\end{algorithm}

After the chains are generated, we estimate the frequency of the patterns in the tail nodes.
Then the confidence of these patterns is calculated, after which the confident patterns are returned as the useful interactions.
We formally describe a basic version of \textit{Random Intersection Chains} in Algorithm \ref{alg:ric}, 
which combines the characteristic of both random intersections and association rule mining.

\begin{algorithm}
  \caption{Random Intersection Chains}
  \label{alg:ric}
  \begin{algorithmic}[1]
  \REQUIRE $\{(X_i, y_i)\}_{i=1}^{N}$(database); \\
  D(length of a chain); \\
  M(number of chains);\\
  $\xi$(threshold of confidence);\\
  \ENSURE $\{L^{(c)}\}_{c\in C}$(returned patterns);
  \STATE{$p^{(c)}\leftarrow |I^{(c)}|/N$ for $c \in C$}
  \STATE{$S^{(c)}\leftarrow \emptyset$, for $c \in C$}
  \FORALL{$c \in C$}
    \STATE{$\{\{S_{d,m}^{(c)}\}_{d=1}^D\}_{m=1}^M\leftarrow$ GenerateChain($\{(X_i, y_i)\}_{i\in I^{(c)}}, D, M$)}
    \STATE{$S^{(c)}\leftarrow \bigcup_{m=1}^M\{s|s\subseteq S_{D,m}^{(c)}\}$}
  \ENDFOR
  \FORALL{$c \in C$}
    \STATE{$L^{(c)}\leftarrow \emptyset$}
    \FORALL{$s \in S^{(c)}$}
      \STATE{$\hat{p}_s^{(c')}\leftarrow$ Frequency($s, \{\{S_{d,m}^{(c')}\}_{d=1}^D\}_{m=1}^M$), for all $c'\in C$}
      \STATE{$\hat{q}_s^{(c)}\leftarrow$ Confidence($\{\hat{p}_s^{(c')}\}_{c'\in C}, \{p^{(c')}\}_{c'\in C}$)}
      \IF{$\hat{q}_s^{(c)}\ge \xi$}
        \STATE{$L^{(c)}\leftarrow L^{(c)}\cup \{s\}$}
      \ENDIF
    \ENDFOR
  \ENDFOR
  \STATE{return $\{L^{(c)}\}_{c\in C}$}
  \end{algorithmic}
\end{algorithm}

We now explain Algorithm \ref{alg:ric} line by line.
There are three parameters in the algorithm, named $D$, $M$ and $\xi$, among which the first two are inherited from random intersections, and the last is from association rule mining.
$D$ represents the length of a chain, $M$ stands for the number of chains and $\xi$ is the threshold of confidence.

Line 1 calculates the proportion of each class, where $I^{(c)}$ represents the indices of observations in class $c$,
$C$ is the set of class labels.
These proportions will be used later to calculate the confidence.
Line 2 initializes the set of frequent patterns, which is now an empty set, for each class.
For every class, $M$ chains are generated at Line 4, then the patterns in the tail nodes as well as all their subpatterns are added to the sets created at Line 2.
The frequency of a pattern in every class is calculated at Line 10, 
after which the confidence is calculated at Line 11.
If the confidence of a pattern is larger than the pre-defined threshold, this pattern will be included in the resulting set at Line 13.
Finally the resulting set is returned at Line 17.

The longer a chain is, the harder for a pattern to appear in its tail node.
Oppositely, the more chains, the more likely for a pattern to be observed in at least one of their tail nodes.
By adjusting $D$ and $M$ carefully, we can control which patterns will be considered as ``frequent''.
As proven in Theorem~\ref{thm:exist_M_D}, there actually exist choices of parameters such that the returned set contains frequent patterns with arbitrarily high probability while including the infrequent ones with probability lower than any given threshold.

\begin{theorem}
  \label{thm:exist_M_D}
  \it Given $\eta_1, \eta_2 \in (0,1]$, for any $\theta \in (0,1]$, there exist choices of $M$, $D$ such that the set $L^{(c)}$ returned by Algorithm \ref{alg:ric} contains $s$ with probability at least $1-\eta_1$ if $P(s\subseteq X|Y=c)\ge \theta$, and with probability at most $\eta_2$ if $P(s\subseteq X|Y=c)< \theta$.
  \hfill
\end{theorem}

From the proof of Algorithm \ref{thm:exist_M_D}, it follows that M and D meet the demand if 
\[\frac{{\rm log}(1-p_1^{D})}{\log(1-p_2^{D})}\ge \frac{a}{b},\]
\[M\ge \frac{a}{\log(1-p_1^{D})},\]
where 
\[ p_1=\min\{p_s:p_s\ge \theta_1\},\]
\[ p_2=\max\{p_s:p_s<\theta_1\},\]
\[ a=\log \eta_1^{-1},\]
\[ b=\log(1-\eta_2)^{-1}.\]

So we have Corollary \ref{crl:choice_M_D}.
\begin{corollary}
  \label{crl:choice_M_D}
  \it M and D meet the requirements in Theorem \ref{thm:exist_M_D} if $M\ge M^*$ and $D\ge D^*$, where
  \begin{equation}
    \label{D}
    D^*=\lceil\max\left\{ \frac{{\rm log}(b+1)-{\rm log}(a)}{{\rm log}(1/p_1)}, \frac{{\rm log}2+{\rm log}a-{\rm log}b}{{\rm log}(1/p_2)-{\rm log}(1/p_1)} \right\}\rceil,
  \end{equation}
  \begin{equation}
    M^*= \lceil \frac{a}{{\rm log}[1/(1-p_1^{D^*})]}\rceil.
  \end{equation}
  \hfill
\end{corollary}

Usually $\eta_1$ and $\eta_2$ are small, thus $a$ is large and $b$ is small.
Assume $a\ge b+1$, then the first item in the braces of Equation~\ref{D} is no greater than 0.
If $a\ge \frac{1}{2}b$, the second item in the braces of Equation~\ref{D} is no less than 0.
In this case, Corollary \ref{crl:choice_M_D_2} holds.

\begin{corollary}
  \label{crl:choice_M_D_2}
  \it If $a\ge \max\left\{b+1, \frac{1}{2}b \right\}$, then M and D meet the requirements in Theorem \ref{thm:exist_M_D} if $M\ge M^*$ and $D\ge D^*$, where
  \begin{equation}
    D^*=\lceil \frac{{\rm log}2+{\rm log}a-{\rm log}b}{{\rm log}(1/p_2)-{\rm log}(1/p_1)}\rceil,
  \end{equation}
  \begin{equation}
    M^*= \lceil \frac{a}{{\rm log}[1/(1-p_1^{D^*})]}\rceil.
  \end{equation}
  \hfill
\end{corollary}

Compared with \textit{Random Intersection Trees}, it is more convenient to apply \textit{Random Intersection Chains} on multi-class classification tasks.
The former is originally designed for binary classification, and detects the interesting patterns for one class at a time.
But it's unable to answer which patterns are the most useful ones among different classes.
On the contrary, since \textit{Random Intersection Chains} not only detects the frequent patterns but also estimates their frequency and confidence, 
we can directly compare the frequency or confidence of patterns in different classes.
Thus we can select the patterns from all the classes simultaneously.

\subsection{Random Intersection Chains with Priority Queues}
\label{sec:ric_queue}
Thanks to its abandon of scanning the complete database, the naive version of \textit{Random Intersection Chains} is more efficient than the traditional methods on large data sets.
Another advantage is that it can extract high-order interactions without discovering all lower-order interactions beforehand.
However, this characteristic is also where a drawback of \textit{Random Intersection Chains} comes from.

It is obvious that any subpattern of a frequent pattern is also frequent.
Since the pattern in the tail node of a chain may contain many components, 
there are in fact a huge number of frequent patterns we have obtained.
For example, if an interaction in a tail node contains $k$ components, 
then all the combinations of these components are frequent.
So there are $2^k-1$ frequent patterns indeed.
It's computationally infeasible to calculate frequency and confidence for every frequent pattern in this sense.
And it's even impossible to pass over all these subpatterns due to its exponential complexity.

\textit{Random Intersection Trees} \citep{shah14} gets over this difficulty by giving priority to the largest frequent patterns.
That is, the authors only consider the patterns whose every superset is infrequent.
They limit their attention to the patterns of the leaf nodes, but ignore their subpatterns.
This may not be a satisfactory solution, since a subpattern is more frequent than the complete pattern, 
and sometimes can also have higher confidence.
Overlooking them may lead to the missing of informative interactions.
Fortunately, by taking advantage of a data structure named priority queue, we find an approach that selects the most frequent patterns from the power set of a given interaction in polynomial time. 

A priority queue is a data structure for maintaining a set of elements, each with an associated value called a key \citep{Cormen09}.
In a priority queue, an element with high priority is served before an element with low priority. 
Like many implementations, we prefer the element enqueued earlier if two elements have the same priority.
This may seem arbitrary, but is meaningful later in this work.
In this paper, we give priority to elements with larger keys, so what we used is exactly a max-priority queue.
A series of operations should be supported by such a queue, and the following are required in our method.
\begin{itemize}
  \item INSERT($S, x, key$): inserts the element $x$ with $key$ into the set $S$, which is equivalent to the operation $S=S\cup \{x\}$;
  \item EXTRACT-MAX($S$): removes and returns the element with the largest key in $S$;
  \item COPY($S$): returns a copy of $S$.
\end{itemize}

We add a new attribute $size$ to a priority queue, which indicates its maximum capacity.
In other words, 
we will discard an element if its $key$ is not the $S.size$ largest, 
which can be done by removing the element with the smallest key after inserting a new element into a full queue.
If $S.size$ is zero, then this priority queue will always be empty.

\begin{algorithm}
  \caption{Random intersection chains with priority queue}
  \label{alg:ric_improved}
  \begin{algorithmic}[1]
    \REQUIRE $\{(X_i, y_i)\}_{i=1}^{N}$(database); \\
    $D$(length of a chain); \\
    $M$(number of chains);\\
    $d_{\rm freq}$(number of frequent patterns);\\
    $d_{\rm conf}$(number of confident patterns);\\
    \ENSURE $\{L^{(c)}\}_{c\in C}$(returned patterns);
    \STATE{$p^{(c)}\leftarrow |I^{(c)}|/N$, for $c \in C$}
    \FORALL{$c \in C$}
      \STATE{$\{\{S_{d,m}^{(c)}\}_{d=1}^D\}_{m=1}^M\leftarrow$ GenerateChain($\{(X_i, y_i=c)\}_{i\in I^{(c)}}, D, M$)}
      \STATE{Initialize $S^{(c)}$ as an empty priority queue of size $d_{\rm freq}$}
      \FOR{$m=1~to~M$}
        \STATE{InsertFreqSubset($S^{(c)},S_{D,m}^{(c)},\{\{S_{d,m}^{(c)}\}_{d=1}^D\}_{m=1}^M$)}
      \ENDFOR
    \ENDFOR
    \FORALL{$c \in C$}
      \STATE{Initialize $L^{(c)}$ as an empty priority queue of size $d_{\rm conf}$}
      \FORALL{$s\in S^{(c)}$}
        \STATE{INSERT($L^{(c)},s,q_s^{(c)}$)}
      \ENDFOR
    \ENDFOR
    \STATE{return $\{L^{(c)}\}_{c\in C}$}
  \end{algorithmic}
\end{algorithm}

With priority queues defined above, we come up with Algorithm \ref{alg:ric_improved}.
The main difference between Algorithm \ref{alg:ric} and Algorithm \ref{alg:ric_improved} lies in the approach of identifying which patterns are frequent or confident.
In Algorithm \ref{alg:ric}, all the patterns in the tail nodes and their subpatterns are considered to be frequent, 
which may result in heavy computation.
On the contrary, Algorithm \ref{alg:ric_improved} only takes the $d_{\rm freq}$ most frequent patterns into consideration.
What's more, Algorithm \ref{alg:ric_improved} returns the $d_{\rm conf}$ most confident patterns, while Algorithm \ref{alg:ric} identifies confident patterns by a pre-defined threshold.

\begin{algorithm}
  \caption{InsertFreqSubset: add the frequent subsets of an itemset}
  \label{alg:insert_freq_subset}
  \begin{algorithmic}[1]
    \REQUIRE $S^{(c)}$ (a priority queue); \\
    $s$(an itemset); \\
    $\{\{S_{d,m}^{(c)}\}_{d=1}^D\}_{m=1}^M$(chains);\\
    \FORALL{$x \in s$}
      \STATE{$\hat{p}_x^{(c)}\leftarrow$ Frequency($\{x\}, \{\{S_{d,m}^{(c)}\}_{d=1}^D\}_{m=1}^M$)}
      \STATE{INSERT($S^{(c)},\{x\},\hat{p}_x^{(c)}$)}
    \ENDFOR
    \FOR{$k=2~to~|s|$}
      \STATE{$A\leftarrow$ COPY($S^{(c)}$)}
      \WHILE{$|A| > 1$}
        \STATE{$a\leftarrow $EXTRACT-MAX($A$)}
        \STATE{$A.size\leftarrow A.size-1$}
        \IF{$|a|=1$ and $a\in S^{(c)}$}
          \STATE{$B\leftarrow$ COPY($A$)}
          \WHILE{$|B| > 0$}
            \STATE{$b\leftarrow $EXTRACT-MAX($B$)}
            \STATE{$B.size\leftarrow B.size-1$}
            \IF{$|b|=k-1$ and $b\in S^{(c)}$ and $a\cap b=\emptyset$}
              \STATE{$\hat{p}_{a\cup b}^{(c)}\leftarrow$ Frequency($a\cup b, \{\{S_{d,m}^{(c)}\}_{d=1}^D\}_{m=1}^M$)}
              \STATE{INSERT($S^{(c)},a\cup b, \hat{p}_{a\cup b}^{(c)}$)}
              \STATE{INSERT($A,a\cup b, \hat{p}_{a\cup b}^{(c)}$)}
              \STATE{INSERT($B,a\cup b, \hat{p}_{a\cup b}^{(c)}$)}
            \ENDIF
          \ENDWHILE
        \ENDIF
      \ENDWHILE
    \ENDFOR
  \end{algorithmic}
\end{algorithm}

The most important improvement of Algorithm \ref{alg:ric_improved} lies in Line 6, where the algorithm named ``InsertFreqSubset'' is used to select the most frequent subsets among the power set of a given itemset.
The algorithm works by selecting frequent itemsets level-wisely.
We have known that an itemset can not be more frequent then any of its subsets.
So if an itemset fails to be in the priority queue, so do its supersets.
For an itemset $s$ consisting of more than one item, it could be uniquely represented by $s=\{a\}\cup (s\setminus \{a\})$, 
where $a$ is the most frequent item in $s$
(if there are several items having the same frequency, choose the one enqueued earliest, which coincides with the implementation of our priority queue).
$\{a\}$ is by definition more frequent than the other singleton subsets of $s\setminus \{a\}$, thus it's more frequent than $s\setminus \{a\}$.
So if we extract the elements from the priority queue in order, $s\setminus \{a\}$ occurs later than $\{a\}$.
This sheds some light on the search for frequent $k$-order itemsets.
We need first take care of the 1-order itemsets in the priority queue.
For each 1-order itemset, we only pay attention to the ($k$-1)-order itemsets that are extracted later than it.
If the 1-order itemset and a ($k$-1)-order itemset are disjoint, then their union is a candidate frequent $k$-order itemset.
We need only calculate the frequency of these $k$-order itemsets.

The detailed InsertFreqSubset algorithm is given in Algorithm \ref{alg:insert_freq_subset}. 
Frequent 1-order itemsets are selected in Line 1-4.
This is followed by a for-loop, where the loop index variable $k$ represents the size of the candidate itemsets.
Since extracting the elements from a queue will change the queue, we make a copy of the queue at Line 5 and Line 11, then apply the EXTRACT-MAX operation on the replica to prevent unwanted changes.
The outer while-loop aims to find 1-order itemsets in the priority queue. 
For an 1-order itemset, the inner while-loop is conducted to find its frequent $k$-order supersets.
($k$-1)-order itemsets in the remaining queue is caught at Line 15,  
then a candidate $k$-order itemset is generated by combining the 1- and ($k$-1)-order itemset, whose frequency is estimated at Line 16.
After that the candidate itemset is inserted into the resulting queue $S^{(c)}$ at Line 17.

The changes of the queue capacity at Line 9 and 14, as well as INSERT operations at Line 18 and 19 have no effects on the resulting queue.
But they can squeeze out the infrequent 1- or ($k$-1)-order itemsets as soon as possible, which reduces the number of candidates and speed up the algorithm.

\section{Computational Complexity}
\label{sec:computation_complexity}
In this section, we analysis the computational complexity of generating a chain and selecting the most frequent subsets of a given itemset, 
from which we can see that the proposed algorithms are very efficient in some sense.

\subsection{Complexity of Chain Generation}
Most intuitively, a chain can be represented by recording each itemsets at a node.
But this not only causes a wastage of space, but also makes it troublesome to generate or check a chain.
For example, to compute the intersection, \cite{shah14} check whether each component of the current interaction is in the new observation.
Every such check is $O(\log p)$ even if a binary search is adopted. 
If most of the components are sufficiently frequent, the size of interactions will keep close to $p$, 
so the time complexity of an intersection can be near $O(p\log p)$.
The total time needed to generate a chain of length $D$ is near $O(Dp\log p)$, and the memory required to store this chain is near $O(Dp)$.

It's worth noting that the itemsets in a chain is descending.
That is to say, the itemset in a node (except the head node) must be a subset of its previous node. 
So rather than record the chain as a series of ordered itemsets, we can view it from the aspect of items.
A chain can be represented by two $p$-dimensional vectors, where the first is a copy of the first randomly chosen observation, 
and the other records how many times the corresponding item occurs in the chain.
For instance, the chain $[C_1=c_1, C_2=c_2, C_3=c_3]\to [C_1=c_1, C_3=c_3]\to [C_1=c_1]$ can be represented by \{$[C_1=c_1, C_2=c_2, C_3=c_3]$, [3, 1, 2]\}, or simply \{$[c_1, c_2, c_3]$, [3, 1, 2]\}.
The memory to store a chain is thus $O(p)$, and is independent of its length.
The memory required to store $M$ chains is the same as $2M$ observations.
For large data sets, the number of observations $N$ can be very huge, thus additional $2M$ observations usually have little influence on the requirement of storage.
An item is in the $i$-th node if and only if it occurs at least $i$ times.
When adding a new node to the chain, we need only pay attention to the items whose number of occurrences equals to the current length of the chain.
If these items also appear in the new randomly chosen observation, then making the intersection can be done simply by adding one to their number of occurrences.
In this way, the time complexity of adding a node is $O(p)$, and generating a chain of length $D$ has a time complexity of $O(pD)$.
Making use of this representation, we have Theorem~\ref{thm:space_complexity_chain} and Theorem~\ref{thm:time_complexity_chain}.

\begin{theorem}
  \label{thm:space_complexity_chain}
  \it The space complexity of Algorithm \ref{alg:generate_chain} is $O(p|C|M)$.
  If $M$ and $D$ are chosen as $M^*$ and $D^*$ in Corollary \ref{crl:choice_M_D_2}, 
  then the return meets the requirements in Theorem \ref{thm:exist_M_D}, 
  while the space complexity is $O(p|C|\lceil \frac{a}{\log[1/(1-p_1^{D^*})]}\rceil)$.
  \hfill
\end{theorem}

\begin{theorem}
  \label{thm:time_complexity_chain}
  \it The time complexity of Algorithm \ref{alg:generate_chain} is $O(p|C|MD)$.
  If $M$ and $D$ are chosen as $M^*$ and $D^*$ in Corollary \ref{crl:choice_M_D_2}, 
  then the return meets the requirements in Theorem \ref{thm:exist_M_D}, 
  while the time complexity is $O(p|C|\lceil \frac{a}{\log[1/(1-p_1^{D^*})]}\rceil D^*)$.\\
  \hfill
\end{theorem}

From Theorem~\ref{thm:space_complexity_chain} and Theorem~\ref{thm:time_complexity_chain} we can conclude that \textit{Random Intersection Chains} will be efficient when $p_1$ is large and $p_2$ is small.
For the ideal situation where $p_1$ approaches 1 and $p_2$ tends to 0, $M^*$ and $D^*$ are near 1, both the space and time complexity of chain generation are linear with the number of features $p$ or the number of different labels in $C$.

\subsection{Complexity of Subset Selection}
As stated earlier in Section~\ref{sec:ric_queue}, finding a frequent pattern $s$ means we have actually found $O(2^{|s|})$ frequent patterns, since every subpattern of $s$ is frequent.
It's not realistic to take all of them into consideration, not even to perform a traversal.
Algorithm \ref{alg:insert_freq_subset} solves this problem with the help of priority queues.
We provide Theorem \ref{thm:freq_subset} to guarantee the validity and time complexity of Algorithm \ref{alg:insert_freq_subset}.
\begin{theorem}
  \label{thm:freq_subset}
  \it If the input priority queue of Algorithm \ref{alg:insert_freq_subset} is empty, then it contains the $d_{\rm freq}$ most frequent subsets of $s$ when the algorithm ends, and the number of frequency calculation is $O(|s|d_{\rm freq}^2)$.
  \hfill
\end{theorem}
If the input priority queue of Algorithm \ref{alg:insert_freq_subset} is not empty, e.g. $S^{(c)}=S'\neq \emptyset$, 
then during the running of the algorithm, itemsets with small frequency in $S'$ will be squeezed out by the subsets of the input itemset $s$, while itemsets with large frequency in $S'$ occupies a position in the priority queue from beginning to end.
So at the end of the algorithm, the priority queue $S^{(c)}$ contains the $d_{\rm freq}$ most frequent itemsets in $S'\cup Pow(s)$.
As a result, after the for-loop in Line 5-7 of Algorithm \ref{alg:ric_improved}, $S^{(c)}$ contains the $d_{\rm freq}$ most frequent itemsets found by the chains for class $c$.


\section{Convergence Analysis}
\label{sec:convergence}
The previous section analyzes the computational complexity of \textit{Random Intersection Chains}, 
from which we can see that it is efficient in some sense.
Another main concern about \textit{Random Intersection Chains} is how effective it is.
In other words, since the frequency and confidence are estimated on the basis of random intersection, 
one may wonder how well they can approximate their true value. 
We find that these estimators have some good properties.
To illustrate this point, asymptotic behaviors of $\hat{p}_s^{(c)}$ and $\hat{q}_s^{(c)}$ are given in Theorem \ref{thm:freq} and Theorem \ref{thm:conf}.
The derivations are given in the appendix, which are mainly based on the multivariate delta method. 

\begin{theorem}
  \label{thm:freq}
  \it $\hat{p}_s^{(c)}$ calculated by Algorithm \ref{alg:frequency} satisfies:
  \begin{equation}
    \sqrt{M}[\hat{p}_s^{(c)}-p_s^{(c)}]\stackrel{d}{\longrightarrow}n(0, \frac{p_s^{(c)}(1-p_s^{(c)})^2}{1-(p_s^{(c)})^D}).
  \end{equation}
  \hfill

\begin{theorem}
  \label{thm:conf}
  \it $\hat{q}_s^{(c)}$ calculated by Algorithm \ref{alg:confidence} satisfies:
  \begin{equation}
    \sqrt{M}[\hat{q}_s^{(c)}-q_s^{(c)}]\stackrel{d}{\longrightarrow} n(0, \tau^2).
  \end{equation}
  where
  \[ \tau^2=\left[\frac{p_s^{(c)}p^{(c)}}{p_s^2}\right]^2\sum_{c'\in C}\frac{[1-p_s^{(c')}]^2p_s^{(c')}}{1-p_s^{(c')D}}p^{(c')2}
  +\left[\frac{p^{(c)}}{p_s^2}\right]^2 \frac{[1-p_s^{(c)}]^2p_s^{(c)}}{1-p_s^{(c)D}}p_s(p_s-2p_s^{(c)}p^{(c)})\]
  \hfill
\end{theorem}
\end{theorem}

From Theorem \ref{thm:freq} we can see that $\hat{p}_s^{(c)}$ converges to an unbiased estimator in distribution as $M$ goes to infinity.
The limiting estimator multiplied by $\sqrt{M}$ would have variance $\frac{p_s^{(c)}(1-p_s^{(c)})^2}{1-(p_s^{(c)})^D}$.
This variance is monotone decreasing with the increase of $D$. 
Remember that the space needed for chains is independent of $D$. 
So if time permits, setting $D$ as large as possible seems a good choice.
The variance tends to 0 if $p_s^{(c)}$ is close to either 0 or 1, 
which means the estimator is more accurate if the itemset is extremely frequent or extremely infrequent.

Theorem \ref{thm:conf} leads to some similar results. 
$\hat{q}_s^{(c)}$ also converges to an unbiased estimator in distribution as $M$ goes to infinity.
But the variance of the limiting estimator multiplied by $\sqrt{M}$ is more complex. 
Anyway, the variance is monotone decreasing with the increase of $D$, too. 
This makes setting larger $D$ more appealing.
In general, large $p_s$(means $s$ is frequent in the whole database), small $p^{(c)}$(means $c$ is a minor class) and extremely large or small $p_s^{(c')}$(means $s$ is either very frequent or very infrequent for each class) will leads to relatively small variance.

\section{Numerical Studies}
To verify the efficiency and effectiveness of \textit{Random Intersection Chains}, we give the results of several numerical examples.
We first conduct a series of experiments on two benchmark data sets for click-through rate (CTR) prediction, which aims to illustrate the efficiency, consistency and effectiveness of \textit{Random Intersection Chains} on large-scale data sets.
We also adopt the data sets used by \cite{shah14}, from whose experimental results two conclusions can be obtained: (1) \textit{Random Intersection Chains} can find almost all meaningful patterns for an ideal data set, (2) rather than act as a classifier themselves, the detected patterns can lead to a better result if they serve as input features for another classification model.
We also show that \textit{Random Intersection Chains} helps to find interactions of numerical features if they are transformed to discrete ones at first, by comparing it to some existing interactive feature selection algorithms on another two UCI data sets.

According to the discussion in Section~\ref{sec:convergence}, longer chains lead to better estimations. 
So we set $D$=100,000, and introduce the maximum order of interaction $K$ as an additional parameter.
A chain stops growing if either its length is larger than $D$, or the number of items in its tail node is no larger than $K$.

\label{sec:experiments}

\subsection{Click-Through Rate Prediction}
Click-through rate (CTR) prediction is an important application of machine learning algorithms, which aims to predict the ratio of clicks to impressions of a specific link. 
The input features associate with either a user or an item, many of which are categorical and of high cardinality.
The label indicates the clicks of a user to an item.
Usually few items will be clicked by a user, which makes the data unbalanced.

We conduct experiments on two public real-world datasets, named Criteo and Avazu.
Criteo data set consists of a portion of Criteo’s traffic over a period of 7 days. There are 45 million users’ clicking records on displayed ads in the data, and the rows are chronologically ordered.
It contains 26 categorical features and 13 numerical features.
Avazu data set contains the records of whether a displayed mobile ad is clicked by a user or not. Click-through data of 10 days, ordered chronologically, is provided. It has 23 features, all of which are categorical. And the total number of samples is above 40 million. 

If one-hot encoding is applied, there will be 998,960 binary features for Criteo data set and 1,544,428 features for Avazu data set, which is obviously unacceptable.
We first unify all the uncommon categories into a single category ``others''.
A category is ``uncommon'' if the number of its occurrences is less than 10 for Criteo or 5 for Avazu.
Then the categorical features are label encoded. 
As for numerical features, a value $z$ will be transformed to $(\log z)^2$ if $z>2$.
Finally, each data set is divided into 10 parts, where 8 parts are used for training, 1 part for validation and 1 part for test.
This procedure is actually the same as in the work of \cite{Song19}, which is also adopted by \cite{Song20, Tsang20}.

\begin{figure}[!t]
  \centering
  \subfigure[Running time of random intersection chains.]{\includegraphics[width=2.7in]{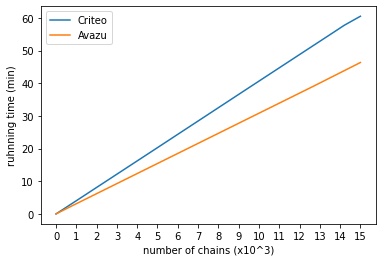}
  \label{running_time}}
  \hfil
  \subfigure[Jaccard-index of the found interactions.]{\includegraphics[width=2.7in]{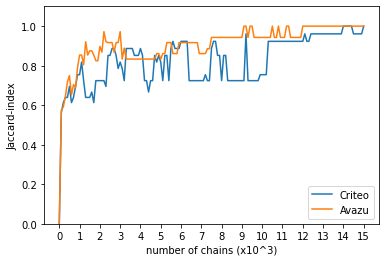}
  \label{Jaccard_index}}
  \caption{Running time and Jaccard-index for Random Intersection Chains with different number of chains on Criteo and Avazu data sets.}
  \label{fig:preformance_m}
\end{figure}

As analyzed in Section~\ref{sec:computation_complexity} and Section~\ref{sec:convergence}, 
the more chains are generated, the more time and memory are needed, but the more accurate estimations of frequency and confidence are obtained.
We apply \textit{Random Intersection Chains} on both data sets with $M$ from 100 to 15,000 for each part in the training set. 
The running time is shown in Figure~\ref{running_time}.
As for memory requirements, the additional space cost caused by \textit{Random Intersection Chains} is at most the same as $2\times (15,000\times 8)=240,000$ observations, which is very small when compared with the original 40 million observations.
To show the consistency of \textit{Random Intersection Chains}, we adopt the Jaccard-index as the criterion of similarity between two sets $S$ and $S'$, which is defined as
\begin{equation}
  {\rm J}(S, S')=\frac{|S\cap S'|}{|S\cup S'|}.
\end{equation}
We calculate the Jaccard-index for interactions found by $M$=15000 and the interactions found by smaller values of $M$, and the results are exhibited in Figure~\ref{Jaccard_index}.
As can be seen from Figure~\ref{running_time}, the running time is linear with the number of chains, and it's relatively small considering the rather large size of the data sets.
Figure~\ref{Jaccard_index} indicates that the returns of \textit{Random Intersection Chains} are very similar for large $M$, which verifies the consistency.

One of the advantages of the interactions defined in this paper is their interpretability.
Since the Avazu data set contains non-anonymized features, we list the 10 most confident interactions with their estimated or accurate frequency and confidence in Table~\ref{tab:interactions}, where we only list out the name of features, and omit the specific value for each feature to keep notation uncluttered.
We can conclude that ``banner\_pos'' plays an important role in advertising.
This observation coincides with the intuition that an advertisement is more likely to be clicked if it's exhibited in a good position.
Many interactions have a feature associated with “app” and a feature about “device”, which indicates the relationship between an item and a user.
Also the estimations of frequency and confidence are pretty good, the RMSE of the estimations are $1.2\times 10^{-3}$, $1.9\times 10^{-3}$ and $1.1\times 10^{-3}$ for Frequency(-), Frequency(-) and Confidence, respectively.
What's more, the corresponding Pearson correlation coefficient are 0.9987, 0.9977 and 0.9879, which means the numerical order of frequency or confidence is well preserved.
Thus the patterns found by \textit{Random Intersection Chains} is likely to be the most frequent and confident ones.

\begin{table}[!t]\scriptsize
  \caption{Ten most confident interactions for Avazu. Frequency(-) stands for the frequency in negative class and Frequency(+) for the frequency in positive class. ``Est.'' represents the values of estimators and ``True'' represents the accurate values in the data sets.}
  \label{tab:interactions}
  \centering
  \begin{tabular}{ccccccc}
    \hline
    \multirow{2}{*}{interaction} & \multicolumn{2}{c}{Frequency(-)} & \multicolumn{2}{c}{Frequency(+)} & \multicolumn{2}{c}{Confidence}\\
    \cline{2-7}
    & Est. & True & Est. & True & Est. & True \\
    \hline
    device\_id, C20 & 0.3607 & 0.3610 & 0.4495 & 0.4518 & 0.2032 & 0.2038 \\
    banner\_pos,app\_domain,app\_category,device\_conn\_type & 0.3646 & 0.3633 & 0.4486 & 0.4497 & 0.2011 & 0.2020 \\
    banner\_pos, app\_category, device\_conn\_type & 0.3646 & 0.3633 & 0.4486 & 0.4497 & 0.2011 & 0.2020 \\
    banner\_pos, app\_domain, app\_category & 0.3835 & 0.3821 & 0.4717 & 0.4735 & 0.2010 & 0.2022 \\
    banner\_pos, app\_category & 0.3835 & 0.3821 & 0.4717 & 0.4735 & 0.2010 & 0.2022 \\
    banner\_pos, app\_id, app\_domain & 0.3759 & 0.3748 & 0.4604 & 0.4619 & 0.2003 & 0.2014 \\
    banner\_pos, app\_id & 0.3759 & 0.3748 & 0.4604 & 0.4619 & 0.2003 & 0.2014 \\
    banner\_pos, app\_domain, device\_conn\_type & 0.3684 & 0.3669 & 0.4498 & 0.4510 & 0.1998 & 0.2009 \\
    device\_type, device\_conn\_type, C20 & 0.3690 & 0.3684 & 0.4502 & 0.4537 & 0.1997 & 0.2012 \\
    banner\_pos, app\_domain & 0.3882 & 0.3867 & 0.4731 & 0.4751 & 0.1995 & 0.2008 \\
    \hline
  \end{tabular}
\end{table}

Finally, the interactions are used as binary features, based on which several popular CTR prediction models are trained.
We adopt 5 models, namely Wide\&Deep \citep{Cheng16}, DeepFM \citep{Guo17}, xDeepFM \citep{Lian18}, Deep\&Cross \citep{Wang17} and AutoInt \citep{Song19}, to test whether adding the interactions to the input is helpful.
The results are shown in Table~\ref{tab:ctr_performance}, where the performance of GLIDER, an interaction detection methods proposed by \cite{Tsang20}, is also presented as a comparison.
We can see that in most cases, adding the interactions found by \textit{Random Intersection Chains} leads to a significant improvement.
In fact, an improvement as small as 0.001 is desirable for the Criteo data set \citep{Cheng16, Guo17, Wang17, Song19, Tsang20}, 
and \textit{Random Intersection Chains} lives up to this expectation.
Perhaps due to the better learning rate we used, our baseline is better than GLIDER.
Despite the difference between baselines, \textit{Random Intersection Chains} makes a comparable improvement to GLIDER.
But according to \cite{Tsang20}, it will take several hours and more than 150 GB memory to perform GLIDER, 
while the requirement of \textit{Random Intersection Chains} is much lower.

\begin{table}
  \caption{CTR prediction performance on two benchmark data sets.
  ``+RIC'' means adding the interactions found by \textit{Random Intersection Chains} to the input. 
  All experiments were repeated for 5 times, and the means are provided with standard deviations in parentheses followed.
  The results which yield a p-value less than 0.05 are shown in bold.
  Rows with * are reported by \cite{Tsang20}.
  ``+GLIDER'' means the inclusion of interactions found by GLIDER. }
  \label{tab:ctr_performance}
  \centering
  \begin{tabular}{lcccc}
    \hline
    \multirow{2}{*}{Model} & \multicolumn{2}{c}{Criteo} & \multicolumn{2}{c}{Avazu}\\
    \cline{2-5}
    & AUC & logloss & AUC & logloss\\
    \hline
    Wide\&Deep & 0.8087(4e-4) & 0.4427(3e-4) & 0.7826(2e-4) & 0.3781(1e-4) \\
    +RIC & \textbf{0.8097(3e-4)} & \textbf{0.4418(2e-4)} & \textbf{0.7828(2e-4)} & \textbf{0.3779(1e-4)} \\
    Wide\&Deep* & 0.8069(5e-4) & 0.4446(4e-4) & 0.7794(3e-4) & 0.3804(2e-4) \\
    +GLIDER* & 0.8080(3e-4) & 0.4436(3e-4) & 0.7795(1e-4) & 0.3802(9e-5) \\
    \hline
    DeepFM & 0.8087(4e-4) & 0.4427(3e-4) & 0.7819(3e-4) & 0.3785(2e-4) \\
    +RIC & \textbf{0.8097(3e-4)} & \textbf{0.4418(2e-4)} & \textbf{0.7826(4e-4)} & \textbf{0.3781(2e-4)} \\
    DeepFM* & 0.8079(3e-4) & 0.4436(2e-4) & 0.7792(3e-4) & 0.3804(9e-5) \\
    +GLIDER* & 0.8097(2e-4) & 0.4420(2e-4) & 0.7795(2e-4) & 0.3802(2e-4) \\
    \hline
    Deep\&Cross & 0.8084(7e-4) & 0.4430(7e-4) & 0.7824(2e-4) & 0.3782(1e-4) \\
    +RIC & \textbf{0.8096(6e-4)} & \textbf{0.4419(5e-4)} & \textbf{0.7835(8e-4)} & \textbf{0.3776(4e-4)} \\
    Deep\&Cross* & 0.8076(2e-4) & 0.4438(2e-4) & 0.7791(2e-4) & 0.3805(1e-4) \\
    +GLIDER* & 0.8086(3e-4) & 0.4428(2e-4) & 0.7792(2e-4) & 0.3803(9e-5) \\
    \hline
    xDeepFM & 0.8082(5e-4) & 0.4432(4e-4) & 0.7824(4e-4) & 0.3782(2e-4) \\
    +RIC & \textbf{0.8103(2e-4)} & \textbf{0.4414(2e-4)} & 0.7825(4e-4) & 0.3781(2e-4) \\
    xDeepFM* & 0.8084(2e-4) & 0.4433(2e-4) & 0.7785(3e-4) & 0.3808(2e-4) \\
    +GLIDER* & 0.8097(3e-4) & 0.4421(3e-4) & 0.7787(4e-4) & 0.3806(1e-4) \\
    \hline
    AutoInt & 0.8077(3e-4) & 0.4436(3e-4) & 0.7788(2e-4) & 0.3804(1e-4) \\
    +RIC & \textbf{0.8090(4e-4)} & \textbf{0.4425(4e-4)} & \textbf{0.7795(3e-4)} & \textbf{0.3802(1e-4)} \\
    AutoInt* & 0.8083 & 0.4434 & 0.7774 & 0.3811 \\
    +GLIDER* & 0.8090(2e-4) & 0.4426(2e-4) & 0.7773(1e-4) & 0.3811(5e-5) \\
    \hline
  \end{tabular}
\end{table}

\subsection{Tic-Tac-Toe Endgame Data}
Tic-Tac-Toe endgame data set \citep{Matheus89, Aha91} encodes the complete set of possible board configurations at the end of tic-tac-toe games.
There are 958 instances and 9 categorical features in the data set.
The possible values for each feature are ``x'', ``o'' and ``b'', which stand for ``black'', ``white'' and ``blank'', respectively. 
There are 8 possible ways to win for both players (3 horizontal lines, 3 vertical lines and 2 diagonal lines).
Our target is to learn these rules that determine which player wins the game. 

This is an ideal data set to test the effectiveness of \textit{Random Intersection Chains}, since all the features are categorical and the label is intrinsically determined by some rules.
As the result of an ending game is completely determined by some rules, 
we can make an accurate prediction if we find all the interesting rules. 
So we make an effort to extract the total 16 valid rules, with different $d_{\rm freq}$.

\begin{figure}[!t]
  \centering
  \includegraphics[width=3.5in]{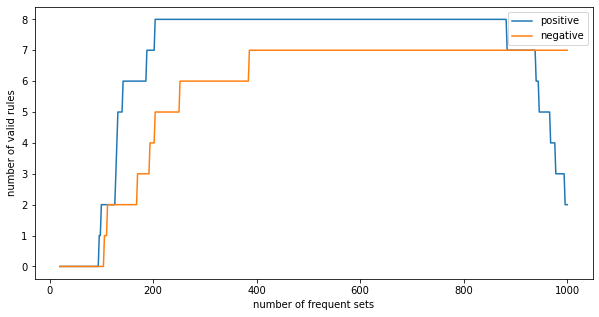}
  \caption{Number of valid rules in confident rules.}
  \label{fig:valid_rule}
\end{figure}

We set $K$=4 and $d_{\rm conf}$=10, so only the 10 most confident patterns for each player are kept.
The range of $d_{\rm freq}$ we adopted is [20, 1000], and the number of found interesting rules is given in Figure~\ref{fig:valid_rule}.
At the beginning, there are more interesting patterns can be found as $d_{\rm freq}$ grows. 
Actually for $d_{\rm freq}$ larger than 400, all the interesting patterns corresponding to x's victory are found, 
while there is one missing pattern for ``o''.
We check the list of the found patterns, and found that the missing pattern is ``a3=o, b3=o, c3=o''.
Its support is 32/958=0.0334, indeed a very small value.
When $d_{\rm freq}$ is too large, not only the missing pattern doesn't occur, but also some already found patterns disappear.
This is because some uncommon patterns have high confidence coincidentally.
For example, ``a1=x, b1=o, b3=b'' occurs 30 times in the database, 
and all the instances containing this pattern happen to be in positive class.
Since the number of the remained patterns are limited, these occasional patterns squeeze out the interesting ones.

The results indicate that the size of priority queue influences the extracted patterns in two ways.
On the one hand, small $d_{\rm freq}$ may lead to a neglect of some meaningful patterns.
On the other hand, if it's too large, some occasionally confident patterns will be chosen, 
which can be treated as the ``overfitting'' phenomenon of \textit{Random Intersection Chains}.

\subsection{Reuters RCV1 Text Data}
The Reuters RCV1 text data contains the tf-idf (term frequency-inverse document frequency) weighted presence of 47,148 word-stems in each document \citep{Lewis04}.

\cite{Lewis04} used a data set consisting of 23,149 documents as the training set.  
Like the processing approach adopted by \cite{shah14}, we only consider word-stems appearing in at least 100 documents and the topics that contain at least 200 documents. 
This leaves 2484 word-stems as predictor variables and 52 topics as prediction targets.
Also, tf-idf is transformed to a binary version, using 1 or 0 to represent whether it is positive.
In this work, we split the original training data into a smaller training set and a test set.
The first 13,149 instances are used for training and the rest for testing.

\begin{figure}[!t]
  \centering
  \subfigure[Precision on the test data.]{\includegraphics[height=6in]{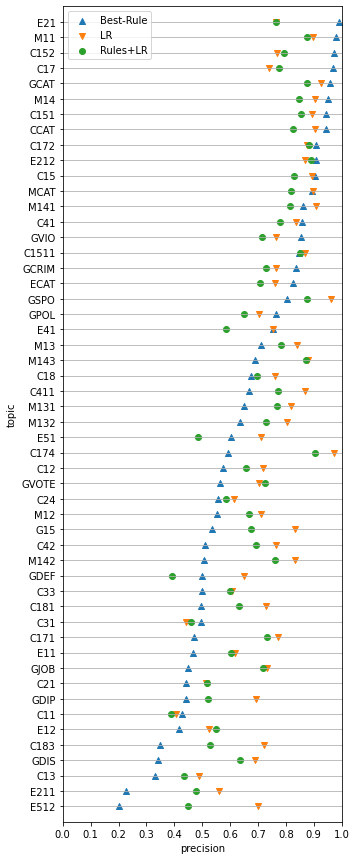}
  \label{precision_rcv1}}
  \hfil
  \subfigure[Recall on the test data.]{\includegraphics[height=6in]{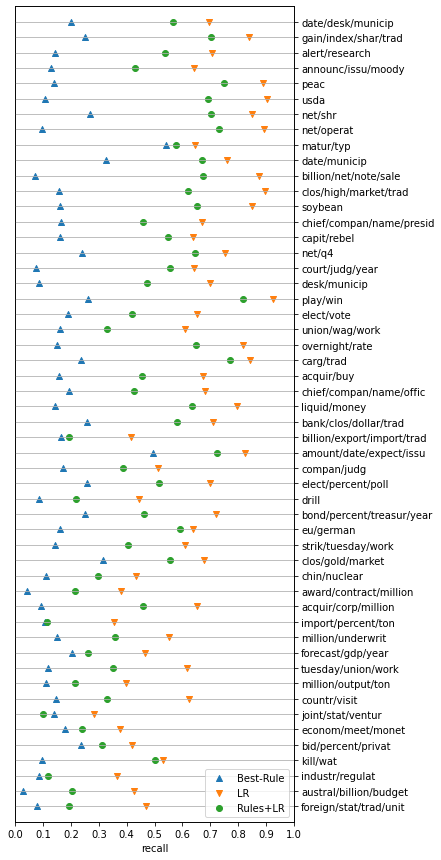}
  \label{recall_rcv1}}
  \caption{Precision and recall on the test data.}
  \label{fig:results_rcv1}
\end{figure}

For each topic $c$, the target is to predict whether a document belongs to this topic.
Setting $d_{\rm freq}$=400, $d_{\rm conf}$=200 and $K$=4, $M$=300, we apply \textit{Random Intersection Chains} on the training set.
Then we evaluate the interactions in two different ways.
The first method, labeled by ``Best-Rule'', is classifying the instances by the best rule directly.
The ``best'' rule is defined as the most confident one among the rules with supports larger than $p^{(c)}$/10, which is the same as what is used by \cite{shah14}.
The other method is treating the rules as selected features for a linear model, which is called ``Rules+LR''.
We also fit a linear model on the total 2484 features as a comparison, labeled as ``LR''.
The precision and recall for the models are given in Figure~\ref{fig:results_rcv1}, where the best rules found by \textit{Random Intersection Chains} are shown in the right side of the figure.

We can see that ``Best-Rule'' sometimes gives good precision or recall, 
but its performance is usually the worst among the three models.
This is not surprising because the information in a single rule is limited, and it's unreasonable to ask a rule to be both general and reliable on a complicated data set.
``Rules+LR'' yields similar precision but generally smaller recall when compared with ``LR''.
At a first glance, it seems \textit{Random Intersection Chains} brings few benefits.
But noticing that the input dimension of the linear model in ``Rules+LR'' is 200, while the number is 2484 in ``LR''.
It rapidly reduces the data size and the computational burden, while the precision and recall only fall slightly.
Also, the interpretability is enhanced since few features are used.

\subsection{Other UCI Data}

We also apply \textit{Random Intersection Chains} on the two data sets used by \cite{shah16}.
The first is Communities and Crime Unnormalized Data Set, ``CCU'' for short, which contains crime statistics for the year 1995 obtained from FBI data, and national census data from 1990.
We take violent crimes per capita as our response, which makes it a regression task.
We process the data in the same way as \cite{shah16}.
This leads to a data set consisting of 1903 observations and 101 features.

The second data set is ``ISOLET'', which consists of 617 features based on the speech waveforms generated from utterances of each letter of the English alphabet.
We consider classification on the notoriously challenging E-set consisting of the letters ``B'', ``C'', ``D'', ``E'', ``G'', ``P'', ``T'', ``V'' and ``Z''.
And finally we have 2700 observations spread equally among 9 classes.

We use the $l_1$-penalised linear regression as the base regression procedure, 
and penalised multinomial regression for the classification example.
The regularization coefficient is determined by 5-fold cross-validation.
To evaluate the procedures, we randomly select 2/3 of the instances for training and the rest for testing. 
This procedure is repeated 200 times for each of the data sets. 
Mean square error is used as the criterion for the regression model and misclassification rate is used for the classification task.
All the settings are exactly the same as those used by \cite{shah16}, 
except we use $l_2$-regularizer to penalise the multinomial regression instead of group Lasso.
This is because we don't know how \cite{shah16} grouped the features.

Since the inputs for these two data sets are numerical and CCU data set corresponds to a regression task, \textit{Random Intersection Chains} can not be applied directly.
To handle this difficulty, the continuous features and response should be transformed to a discrete version.
The response of CCU data set is split into 5 categories by quantiles, and all the continuous features are then split to 5 intervals according to information gain. 
Setting $d_{\rm freq}$=2500, $d_{\rm conf}$=1225 and $K=5$, $M=300$, we add the product $X_{i_1}X_{i_2}\cdots X_{i_k}$ as an interactive feature to the input if there is a rule in the form of ($X_{i_1}$=$x_{i_1}$, $X_{i_2}$=$x_{i_2}$, ..., $X_{i_k}$=$x_{i_k}$) for some $x_{i_1}$, $x_{i_2}$, ..., $x_{i_k}$ in the resulting rule sets.
The results of models with and without adding the rules found by \textit{Random Intersection Chains} are shown in Table \ref{tab:ccu_isolet}, labeled by ``RIC'' and ``Main''.
We also list out the results reported by \cite{shah16}, including base procedures (``Main*''), iterated Lasso fits (``Iterated''), Lasso following marginal screening for interactions (``Screening''), Backtracking, Random Forests \citep{Breiman01}, hierNet \citep{bien13} and MARS \citep{Friedman91}.
For CCU data set, our base model outperforms the one used by \cite{shah16}, which may caused by a better penalty parameter.
\textit{Random Intersection Chains} leads to comparable or better result when compared to existing algorithms. 
As for ISOLET data set, the result of our base model is not as good as the one used by \cite{shah16}.
This is not surprising since we simply use $l_2$-regularizer while \cite{shah16} adopted group Lasso to penalise the model.
But we can see that \textit{Random Intersection Chains} can run on this data set and leads to a good improvement, while some existing methods such as Screening, hierNet, MARS are inapplicable.
We think this could be an evidence of our method's efficiency.

\begin{table}[!t]
  \caption{Results of CCU and ISOLET. Rows with * are reported by \cite{shah16}.}
  \label{tab:ccu_isolet}
  \centering
  \begin{tabular}{c||c|c} 
      \hline
      \multirow{2}{*}{method} & \multicolumn{2}{c}{ERROR}\\
      \cline{2-3}
      &  Communities and crime & ISOLET\\
      \hline
      Main & $0.404(5.7\times 10^{-3})$ & $0.0730(5.5\times 10^{-4}$) \\
      RIC & $0.369(6.3\times 10^{-3}$) & $0.0665(5.3\times 10^{-4}$) \\
      Main* & $0.414(6.5\times 10^{-3}$) & $0.0641(4.7\times 10^{-4}$)\\
      Iterate* & $0.384(5.9\times 10^{-3}$) & $0.0641(4.7\times 10^{-4}$)\\
      Screening* & $0.390(7.8\times 10^{-3}$) & - \\
      Backtracking* & $0.365(3.7\times 10^{-3}$) & $0.0563(4.5\times 10^{-4}$)\\
      Random Forest* & $0.356(2.4\times 10^{-3}$) & $0.0837(6.0\times 10^{-4}$)\\
      hierNet* & $0.373(4.7\times 10^{-3}$) & - \\
      MARS* & $5580.586(3.1\times 10^{3}$) & - \\
      \hline
  \end{tabular}
\end{table}

\section{Conclusion}
\label{sec:conclusion}
Based on the contents of association rule mining and random intersections, we propose \textit{Random Intersection Chains} to discover meaningful categorical interactions.
A number of chains are generated by intersections, and the patterns in the tail nodes are regarded as frequent patterns.
Then the frequency and confidence of these patterns are estimated by maximum likelihood estimation and Bayes’ theorem.
Finally the most confident patterns are selected.
An efficient algorithm for selecting the most frequent subpatterns from a given pattern in polynomial time is also provided.

We prove that there exist appropriate parameters that can keep all the frequent patterns, while the infrequent ones are prevented. 
The time and space complexities are analyzed, showing that the algorithm is both time- and memory-efficient.
The asymptotic behavior of the estimations is guaranteed.
When the number of chains goes to infinity, the estimated frequency and confidence converge to their true values.

As a supplementary, s series of experiments are conducted to verify the effectiveness of \textit{Random Intersection Chains}.
We show it's time efficient and consistent to detect the most frequent and confident patterns by applying the algorithm on two CTR prediction data sets.
The prediction result verifies that adding these interactions are beneficial for CTR prediction.
The ability of detecting useful patterns is further tested on the Tic-Toc-Toe data, where almost all the meaningful rules are found if parameters are appropriately chosen.
The experiments on Reuters RCV1 Text data show that the found patterns can not only serve as a classifier themselves, but also be the input features for another model.
We also compare our algorithm with some other interaction detection methods on several UCI data sets with continuous features or response.
The results show that \textit{Random Intersection Chains} can help if the features or response are transformed into categorical ones beforehand.

One limitation of \textit{Random Intersection Chains} is that it can not be applied directly on numerical features or response.
We are trying to extend its application domain to these cases.
Another difficulty lies in the choice of parameters.
Different parameter settings influences the prediction performance, but tuning the parameters by grid search is time-consuming.
We hope to find a better approach to chose the parameters.


\acks{This work was partially supported by the National Natural Science Foundation of China under grants 11671418 and 12071428 and by the Zhejiang Provincial Natural Science Foundation of China under grant LZ20A010002.}


\newpage

\appendix
\section*{Appendix A.}
\label{app:theorem}



In this appendix we give the proofs omitted earlier in the paper.

\noindent
{\bf Proof of Theorem \ref{thm:exist_M_D}}. 
To keep the notation uncluttered, we omit the superscript ``$(c)$'' on the probabilities.
For a pattern $s$, we use the notation $p_s$ for the probability of $s$'s occurrence conditioned on $Y=c$.
And define
\[ p_1=\min\{p_s:p_s\ge \theta\},\]
\[ p_2=\max\{p_s:p_s<\theta\}.\]
For a chain of length $D$, 
\[\mathbb{P}(s\subseteq S_{D,1}^{(c)})=p_s^D. \]
And for $M$ chains,
\[\mathbb{P}(s\subseteq S_{D,M}^{(c)})=1-[1-\mathbb{P}(s\subseteq S_{D,1}^{(c)})]^M=1-[1-p_s^D]^M\eqqcolon g(p_s;D,M).\]
We can see that $g(p_s;M,D)$ is monotone increasing with the increasing of $p_s$ and $M$, and the decreasing of $D$.
For $p_s\ge \theta$, if $M\ge \frac{{\log}\eta_1}{\log(1-p_1^D)}$, then 
\begin{equation}
  \begin{aligned}
    \mathbb{P}(S\subseteq S_{D,M}^{(c)})&=g(p_s;D,M)\\
    &\ge g(p_1;D,\frac{{\log}\eta_1}{\log(1-p_1^D)})\\
    &=1-[1-p_1^D]^{\frac{{\log}\eta_1}{\log(1-p_1^D)}}\\
    &=1-\eta_1^{\frac{{\log}(1-p_1^D)}{{\log}(1-p_1^D)}}\\
    &=1-\eta_1.
    \nonumber
  \end{aligned}
\end{equation}
Define 
\[M^*(D)=\lceil \frac{\log\eta_1}{\log(1-p_1^D)}\rceil, \]
\[\bar{M}(D)=\frac{\log\eta_1}{\log(1-p_1^D)}+1. \]
Thus $\bar{M}(D)\ge M^*(D)\ge \frac{\log\eta_1}{\log(1-p_1^D)}$. Then for $p_s\ge \theta$, we have $\mathbb{P}(S\subseteq S_{D,M}^{(c)})\ge 1-\eta_1$ if $M\ge M^*(D)$.

Next we give the conditions for $S^{(c)}$ containing $s$ with probability at most $\eta_2$ if $P(s\subseteq X|Y=c)< \theta$.
Fixing $M=M^*(D)$, for $p_s< \theta$ we have
\begin{equation}
  \begin{aligned}
    \mathbb{P}(S\subseteq S_{D,M^*}^{(c)})&=g(p_s;D,M^*)\\
    &< g(p_2;D,\bar{M})\\
    &=1-[1-p_2^D]^{\frac{{\log}(\eta_1)}{\log(1-p_1^D)}+1}\\
    &=1-\eta_1^{\frac{{\log}(1-p_2^D)}{{\log}(1-p_1^D)}}(1-p_2^D).
  \end{aligned}
  \nonumber
\end{equation}
Define
\[f(D)=\frac{{\log}(1-p_2^D)}{{\log}(1-p_1^D)}.\]
Take the derivative of $f$, then we have
\begin{equation}
  \begin{aligned}
    f'(D)&= \frac{-p_2^D\log p_2}{1-p_2^D}\log(1-p_1^D)+\frac{p_1^D\log p_1}{1-p_1^D}\log(1-p_2^D)\\
    &= \log(1-p_1^D)\log(1-p_2^D)[f_1(p_1^D)-f_1(p_2^D)],
  \end{aligned}
  \nonumber
\end{equation}
where 
\[f_1(x)= \frac{x\log x}{(1-x)\log(1-x)}.\]
So the corresponding derivative is
\[f_1'(x)= \frac{(1+\log x-x)\log(1-x)+x\log x}{[(1-x)\log(1-x)]^2}.\]
Denote the numerator as $f_2(x)$, and take the derivative, then we have
\[f_2(x)=(1+\log x-x)\log(1-x)+x\log x,\]
\[f_2'(x)= \frac{(1-x)^2\log(1-x)-x^2\log x}{x(1-x)}.\]
Again denoting the numerator as $f_3(x)$ and taking the derivative, we have
\[f_3(x)=(1-x)^2\log(1-x)-x^2\log x,\]
\[f_3'(x)=-2(1-x)\log(1-x)-2x\log x-1.\]
Denoting $f_4(x)=f_3'(x)$, we have
\[f_4'(x)=2\log(1-x)-2\log x=2\log(\frac{1}{x}-1). \]
Therefore for $x\in (0,1)$,
\begin{equation}
  \begin{aligned}
    &f_4'(x)<0~{\rm for}~x\in (0, \frac{1}{2}),~f_4'(x)>0~{\rm for}~x\in (\frac{1}{2}, 1)\\
    \Rightarrow &f_3'(x)= f_4(x)\le f_4(\frac{1}{2})=-1<0\\
    \Rightarrow &f_3(x)\le \lim_{x\to 0}f_3(x)=0\\
    \Rightarrow &f_2'(x)\le 0\\
    \Rightarrow &f_2(x)\le \lim_{x\to 0}f_2(x)=0\\
    \Rightarrow &f_1'(x)\le 0.
  \end{aligned}
  \nonumber
\end{equation}
Noticing that $0\le p_2<p_1\le 1$, we have $f_1(p_1^D)<f_1(p_2^D)$, and thus $f'(D)<0$.
So $g(p_2;D,\bar{M})$ is a monotone decreasing function of $D$.
Extend the domain of $f$ to real numbers, according to $L'H\hat{o}pital's$ rule,
\begin{equation}
  \begin{aligned}
    \lim_{x\to \infty}f(x)&=\lim_{x\to \infty}{(\frac{-p_2^x\log p_2}{1-p_2^x}/\frac{-p_1^x\log p_1}{1-p_1^x})}\\
    &=\frac{\log p_2}{\log p_1} \lim_{x\to \infty}(\frac{p_2}{p_1})^x \lim_{x\to \infty}\frac{1-p_1^x}{1-p_2^x}\\
    &=\frac{\log p_2}{\log p_1}\cdot 0 \cdot 1 = 0.
  \end{aligned}
\end{equation}
Then according to Heine theorem, 
\[\lim_{D\to \infty}f(D)=0.\]
Thus we have
\[\lim_{D\to \infty} g(p_2;D,\bar{M})=1-\lim_{D\to \infty}\eta_1^{f(D)}(1-p_2^D)=0.\]
So for any $\eta_2\in (0,1)$, there exists $D^*\in \mathbb{N}$ such that $\mathbb{P}(S\subseteq S_{D,M^*}^{(c)})\le \eta_2$ if $D\ge D^*$. 
\hfill\BlackBox

~\\
\noindent
{\bf Proof of Corollary \ref{crl:choice_M_D}}.
From the proof of Theorem \ref{thm:exist_M_D}, it follows that there exist a choice $M^*$ and $D^*$, where $D^*$ is a feasible solution of $D$ subject to
\begin{equation}
  \begin{aligned}
    &1-[1-p_2^D]^{\frac{{\log}(\eta_1)}{\log(1-p_1^D)}+1}\le \eta_2\\
    \Leftrightarrow &(1-p_2^D)\eta_1^{\frac{\log(1-p_2^D)}{\log(1-p_1^D)}}\ge 1-\eta_2\\
    \Leftrightarrow &\frac{\log \eta_1}{\log(1-p_1^D)}\le \frac{\log(1-\eta_2)}{\log(1-p_2^D)}-1.
  \end{aligned}
  \nonumber
\end{equation}
The inequality $\frac{x}{1+x}\le \log(1+x)\le x$ holds for $x>-1$.
Substituting $\frac{p_1^D}{1-p_1^D}$ for $x$ in the first inequality and rearranging, we have
\begin{equation}
\label{eq:ineq1}
    \frac{1}{\log(1-p_1^D)^{-1}}\le \frac{1}{p_1^D}.
\end{equation}
Similarly taking the place of $x$ by $\frac{p_2^D}{1-p_2^D}$ in the second inequality and rearranging, we have
\begin{equation}
\label{eq:ineq2}
    \frac{1}{p_2^D}-1\le \frac{1}{\log(1-p_2^D)^{-1}}.
\end{equation}
Taking advantage of Inequalities \ref{eq:ineq1} and \ref{eq:ineq2}, if
\begin{equation}
  \label{eq:condition_d}
  \frac{\log \eta_1^{-1}}{p_1^D}\le (\frac{1}{p_2^D}-1)\log(1-\eta_2)^{-1}-1,
\end{equation}
then
\[\frac{\log \eta_1^{-1}}{\log(1-p_1^D)^{-1}}\le \frac{\log \eta_1^{-1}}{p_1^D}\le (\frac{1}{p_2^D}-1)\log(1-\eta_2)^{-1}-1\le \frac{\log(1-\eta_2)^{-1}}{\log(1-p_2^D)^{-1}}-1, \]
where the first, second and third inequality is obtained from Inequalities \ref{eq:ineq1}, \ref{eq:condition_d} and \ref{eq:ineq2}, respectively.
So we need only to ensure Inequality \ref{eq:condition_d}.
Denote $a=\log \eta_1^{-1}, b=\log(1-\eta_2)^{-1}$,

\begin{equation}
  \begin{aligned}
    {\rm Inequality}~\ref{eq:condition_d}&\Leftrightarrow ap_1^{-D}\le b(p_2^{-D}-1)-1\\
    &\Leftrightarrow ap_1^{-D}\left[\frac{b}{a}\left(\frac{p_1}{p_2}\right)^D-1\right]\ge b+1 \\
    &\Leftarrow \begin{cases}
                  ap_1^{-D}\ge b+1 \\
                  \frac{b}{a}\left(\frac{p_1}{p_2}\right)^D\ge 2
                \end{cases}\\
    &\Leftarrow \begin{cases}
                  D\ge \frac{\log(b+1)-log(a)}{\log p_1^{-1}}\eqqcolon D_1 \\
                  D\ge \frac{\log 2+\log a-\log b}{\log p_2^{-1}-\log p_1^{-1}}\eqqcolon D_2
                \end{cases}.\\
  \end{aligned}
  \nonumber
\end{equation}
So $D^*=\lceil\max\{D_1, D_2\}\rceil, M^*=\lceil \frac{a}{\log(1-p_1^{D^*})}\rceil$ are subject to the conditions in Theorem \ref{thm:exist_M_D}.
\hfill\BlackBox

~\\
\noindent
{\bf Proof of Theorem \ref{thm:space_complexity_chain}}.
As stated in Section~\ref{sec:computation_complexity}, we can represent a chain by two vectors of dimension $p$.
Thus for every chain, the memory needed is $2p$.
For every class in $C$, $M$ chains are generated.
Thus the entire space complexity is therefore $2p|C|M$.
\hfill\BlackBox

~\\
\noindent
{\bf Proof of Theorem \ref{thm:time_complexity_chain}}.
For the $d$-th node of the $m$-th chain, we need to check whether the element in $S_{d-1, m}$ occurs in $X_{i_d}$, 
the time complexity is $p$. 
So the time complexity of generating a chain is $O(pD)$.
And the total time complexity is $O(p|C|MD)$.
\hfill\BlackBox

~\\
\noindent
{\bf Proof of Theorem \ref{thm:freq_subset}}.
Denote 
\[S_k=\{s':|s'|\le k, s'\subseteq s\},\]
\[F_k=\{{\rm the}~d_{\rm freq}~{\rm most~frequent~patterns~in}~S_k\}.\]
We prove the first part of this theorem by the loop invariant: 
\emph{at the beginning of the $i$th iteration of Line 5-24,
the priority queue $S^{(c)}$ contains the patterns in $F_k$.}

{\bf Initialization}: Before the loop in Line 5-24, this loop invariant is true since every element in $S$ was inserted into $S^{(c)}$ in Line 1-4.
By the property of priority queue, $S^{(c)}$ stores the $d_{\rm freq}$ most frequent elements, each as an 1-order pattern.

{\bf Maintenance}: During the $i$-th iteration, by definition we have $(F_{i+1}\setminus F_i)\subseteq S_{i+1}$ and $(F_{i+1}\cap S_i)\subseteq F_i$.
If $F_{i+1}=F_i$, no patterns can be inserted into the priority queue, thus $S^{(c)}$ keeps unchanged, and satisfies the loop invariant.
Otherwise, for any pattern $s'\in (F_{i+1}\setminus F_i)$, we can write $s'=a\cup (s'\setminus a)$, where $a$ is the most frequent 1-order subpattern of $s'$. 
Since $(s'\setminus a)$ is more frequent than $s'$, it belongs to $F_{i+1}$. 
Noticing $(s'\setminus a)$ is an $i$-order pattern, it belongs $S_i$, thus it also belongs to $(F_{i+1}\cap S_i)\subseteq F_i$.
By the definition of $a$, its frequency is larger than any 1-order subpattern of $(s'\setminus a)$, and thus also larger than $(s'\setminus a)$.
According to the property of priority queue, patterns with larger frequency are extracted earlier.
So when $a$ is extracted from the priority queue $A$ at Line 8, $(s'\setminus a)$ is still in $A$.
And $(s'\setminus a)$ will be extracted from the queue $B$ later at Line 13.
$a$ and $(s'\setminus a)$ satisfy the conditions in Line 14 and thus $s'$ will be inserted into $S^{(c)}$ at Line 17.
Since every pattern in $F_{i+1}\setminus F_i$ will not be missed by the algorithm, and $S^{(c)}$ contains all the patterns in $F_i$ at the beginning of iteration by the assumption, we can conclude that $S^{(c)}$ contains every pattern in $F_{i+1}$.
The loop invariant is still true.

{\bf Termination}: When the loop ends, it is the $|s|$-th iteration, thus $S^{(c)}$ contains all patterns in $F_{|s|}$, which is exactly what we need.

As for computational complexity, we only need to calculate the frequency in Line 2 and 16. It can be seen directly that Line 2 repeats $|s|$ times. Line 16 repeats at most $d_{\rm freq}$ times due to Line 12, which will as a whole repeats at most $d_{\rm freq}$ times due to Line 7, and finally repeats $|s|$ times due to Line 5 in total. Thus there are at most $O(|s|d_{\rm freq}^2)$ computations of frequency overall.
\hfill\BlackBox

~\\
To prove Theorem \ref{thm:freq} and \ref{thm:conf}, we need to to know the expectation and variance of $k_{s,m}^{(c)}$, $\chi_{s,m}^{(c)}$ and their covariance. So we give Lemma \ref{lemma:distrib_k}, \ref{lemma:distrib_chi} and \ref{lemma:cov_kchi} first.
To simplify the notation, we omit the superscript ``$(c)$'' unless otherwise stated.
\begin{lemma}
  \label{lemma:distrib_k}
  \it The expectation and variance of $k_{m,s}$ in Algorithm \ref{alg:frequency} is 
  \[{\rm E}[k_{m,s}]=\frac{p_s(1-p_s^D)}{1-p_s},\]
  \[{\rm Var}[k_{m,s}]=\frac{p_s-(2D+1)p_s^{D+1}+(2D+1)p_s^{D+2}-p_s^{2D+2}}{(1-p_s)^2}.\]
  \hfill
\end{lemma}
{\bf Proof}.
\begin{equation}
  \begin{aligned}
    {\rm E}[k_{m,s}]&=\sum_{d=0}^D{\mathbb{P}(k_{m,s}=d)\cdot d}\\
    &=\sum_{d=0}^{D-1}p_s^d(1-p_s)d+p_s^DD\\
    &=\frac{p_s-Dp_s^D+(D-1)p_s^{D+1}}{1-p_s}+Dp_s^D\\
    &=\frac{p_s(1-p_s^D)}{1-p_s},
  \end{aligned}
  \nonumber
\end{equation}
\begin{equation}
  \begin{aligned}
    {\rm E}[k_{m,s}^2]&=\sum_{d=0}^D{\mathbb{P}(k_{m,s}=d)\cdot d^2}\\
    &=\sum_{d=0}^{D-1}p_s^d(1-p_s)d^2+p_s^DD^2\\
    &=\frac{p_s+p_s^2-(2D+1)p_s^{D+1}+(2D-1)p_s^{D+2}}{(1-p_s)^2},
  \end{aligned}
  \nonumber
\end{equation}
\begin{equation}
  \begin{aligned}
    {\rm Var}[k_{m,s}]&={\rm E}[k_{m,s}^2]-({\rm E}[k_{m,s}])^2\\
    &=\frac{p_s-(2D+1)p_s^{D+1}+(2D+1)p_s^{D+2}-p_s^{2D+2}}{(1-p_s)^2}.
  \end{aligned}
  \nonumber
\end{equation}
\hfill\BlackBox
~\\
\begin{lemma}
  \label{lemma:distrib_chi}
  \it The expectation and variance of $\chi_{m,s}$ in Algorithm \ref{alg:frequency} is 
  \[{\rm E}[\chi_{m,s}]=1-p_x^D,\]
  \[{\rm Var}[\chi_{m,s}]=p_x^D(1-p_x^D).\]
  \hfill
\end{lemma}
{\bf Proof}.
\[{\rm E}[\chi_{m,s}]={\rm E}[\chi_{m,s}^2]=\mathbb{P}(k_{m,s}<D)=1-p_x^D,
\]
\[{\rm Var}[\chi_{m,s}]={\rm E}[\chi_{m,s}^2]-({\rm E}[\chi_{m,s}])^2=p_x^D(1-p_x^D).\]
\hfill\BlackBox
~\\
\begin{lemma}
  \label{lemma:cov_kchi}
  \it The covariance of $k_{m,s}$ and $\chi_{m,s}$ in Algorithm \ref{alg:frequency} is 
  \[{\rm Cov}[k_{m,s}, \chi_{m,s}]=\frac{-Dp_s^D+(D+1)p_s^{D+1}-p_s^{2D+1}}{1-p_s}.
  \]
  \hfill
\end{lemma}
{\bf Proof}.
\[{\rm E}[k_{m,s}\chi_{m,s}]=\sum_{d=0}^{D-1}p_s^d(1-p_s)d=\frac{p_s-Dp_s^D+(D-1)p_s^{D+1}}{1-p_s},
\]
\[{\rm Cov}[k_{m,s},\chi_{m,s}]={\rm E}[k_{m,s}\chi_{m,s}]-{\rm E}[k_{m,s}]{\rm E}[\chi_{m,s}]=\frac{-Dp_s^D+(D+1)p_s^{D+1}-p_s^{2D+1}}{1-p_s}.\]
\hfill\BlackBox
~\\
\noindent
{\bf Proof of Theorem \ref{thm:freq}}.
We need to determine the distribution of $K_{m,s}^{(c)}$ and $I_{m,s}^{(c)}$.

Define 
\[g(k,\chi)=\frac{k}{k+\chi},\]
\[g_k'\coloneqq \left.\frac{\partial g}{\partial k}\right|_{{\rm E}k_{m,s},{\rm E}\chi_{m,s}}=\frac{(1-p_s)^2}{1-p_s^D}, \]
\[g_\chi'\coloneqq \left.\frac{\partial g}{\partial \chi}\right|_{{\rm E}k_{m,s},{\rm E}\chi_{m,s}}=\frac{-p_s(1-p_s)}{1-p_s^D}, \]
\[\tau^2\coloneqq (g_k')^2{\rm Var}[k_{m,s}]+(g_\chi')^2{\rm Var}[\chi_{m,s}]+2g_k'g_\chi'{\rm Cov}[k_{m,s},\chi_{m,s}]=\frac{p_s(1-p_s^2)}{1-p_s^D}. \] 
By the Multivariate Delta Method, 
\[\sqrt{M}(\hat{p}_s-p_s)=\sqrt{M}(g(\bar{k}_s, \bar{\chi}_s)-p_s)\to n(0,\tau^2) \]
in distribution.
\hfill\BlackBox

~\\
\noindent
{\bf Proof of Theorem \ref{thm:conf}}.
Without loss of generality, we take $\hat{q}_s^{(1)}$ for example.
From the definition of $\hat{q}_s^{(1)}$, we have
\[
  \hat{q}_s^{(1)}=\frac{\hat{p}_s^{(1)}p^{(1)}}{\sum_{c\in C}\hat{p}_s^{(c)}p^{(c)}}
  =\frac{\frac{\bar{k}_s^{(1)}}{\bar{k}_s^{(1)}+\bar{\chi}_s^{(1)}}p^{(1)}}{\sum_{c\in C}\frac{\bar{k}_s^{(c)}}{\bar{k}_s^{(c)}+\bar{\chi}_s^{(c)}}p^{(c)}}
  \eqqcolon h(\{\bar{k}_s^{(c)}, \bar{\chi}_s^{(c)}\}_{c\in C}),
\]
The partial derivatives are
\begin{equation}
  \begin{aligned}
    \frac{\partial h}{\partial \bar{k}_s^{(c)}}&=
    \begin{cases}
      \frac{\frac{\bar{\chi}_s^{(1)}}{\left(\bar{k}_s^{(1)}+\bar{\chi}_s^{(1)}\right)^2}p^{(1)}\sum_{c\in C}\frac{\bar{k}_s^{(c)}}{\bar{k}_s^{(c)}+\bar{\chi}_s^{(c)}}p^{(c)}-\frac{\bar{\chi}_s^{(1)}}{\left(\bar{k}_s^{(1)}+\bar{\chi}_s^{(1)}\right)^2}p^{(1)}\frac{\bar{k}_s^{(1)}}{\bar{k}_s^{(1)}+\bar{\chi}_s^{(1)}}p^{(1)}}{\left(\sum_{c\in C}\frac{\bar{k}_s^{(c)}}{\bar{k}_s^{(c)}+\bar{\chi}_s^{(c)}}p^{(c)}\right)^2}, \hfil \mbox{if~} c=1\\
      \frac{-\frac{\bar{\chi}_s^{(c)}}{\left(\bar{k}_s^{(c)}+\bar{\chi}_s^{(c)}\right)^2}p^{(c)}\frac{\bar{k}_s^{(1)}}{\bar{k}_s^{(1)}+\bar{\chi}_s^{(1)}}p^{(1)}}{\left(\sum_{c\in C}\frac{\bar{k}_s^{(c)}}{\bar{k}_s^{(c)}+\bar{\chi}_s^{(c)}}p^{(c)}\right)^2}, \hfill \mbox{if~} c\neq 1\\
    \end{cases},\\
    \frac{\partial h}{\partial \bar{\chi}_s^{(c)}}&=
    \begin{cases}
      \frac{\frac{-\bar{k}_s^{(1)}}{\left(\bar{k}_s^{(1)}+\bar{\chi}_s^{(1)}\right)^2}p^{(1)}\sum_{c\in C}\frac{\bar{k}_s^{(c)}}{\bar{k}_s^{(c)}+\bar{\chi}_s^{(c)}}p^{(c)}+\frac{\bar{k}_s^{(1)}}{\left(\bar{k}_s^{(1)}+\bar{\chi}_s^{(1)}\right)^2}p^{(1)}\frac{\bar{k}_s^{(1)}}{\bar{k}_s^{(1)}+\bar{\chi}_s^{(1)}}p^{(1)}}{\left(\sum_{c\in C}\frac{\bar{k}_s^{(c)}}{\bar{k}_s^{(c)}+\bar{\chi}_s^{(c)}}p^{(c)}\right)^2}, \hfil \mbox{if~} c=1\\
      \frac{\frac{\bar{k}_s^{(c)}}{\left(\bar{k}_s^{(c)}+\bar{\chi}_s^{(c)}\right)^2}p^{(c)}\frac{\bar{k}_s^{(1)}}{\bar{k}_s^{(1)}+\bar{\chi}_s^{(1)}}p^{(1)}}{\left(\sum_{c\in C}\frac{\bar{k}_s^{(c)}}{\bar{k}_s^{(c)}+\bar{\chi}_s^{(c)}}p^{(c)}\right)^2}, \hfill \mbox{if~} c\neq 1
    \end{cases}.
  \end{aligned}
  \nonumber
\end{equation}

Substitute the variables with their expectations, and denote $p_s=\sum_{c\in C}p_s^{(c)}p^{(c)}$ as the marginal probability of $s$, then
\begin{equation}
  \begin{aligned}
    h_{\bar{k}_s^{(c)}}'\coloneqq \frac{\partial h}{\partial \bar{k}_s^{(c)}}(\mu)&=
    \begin{cases}
      \frac{1}{p_s^2}\left[\frac{(1-p_s^{(1)})^2}{1-p_s^{(1)D}}p^{(1)}p_s-\frac{(1-p_s^{(1)})^2}{1-p_s^{(1)D}}p^{(1)}p_s^{(1)}p^{(1)}\right], \hfil \mbox{if~} c=1\\
      \frac{1}{p_s^2}\left[-\frac{(1-p_s^{(c)})^2}{1-p_s^{(c)D}}p^{(c)}p_s^{(1)}p^{(1)}\right], \hfill \mbox{if~} c\neq 1\\
    \end{cases},\\
    h_{\bar{\chi}_s^{(c)}}'\coloneqq \frac{\partial h}{\partial \bar{\chi}_s^{(c)}}(\mu)&=
    \begin{cases}
      \frac{1}{p_s^2}\left[-\frac{(1-p_s^{(1)})p_s^{(1)}}{1-p_s^{(1)D}}p^{(1)}p_s+\frac{(1-p_s^{(1)})p_s^{(1)}}{1-p_s^{(1)D}}p^{(1)}p_s^{(1)}p^{(1)}\right], \hfil \mbox{if~} c=1\\
      \frac{1}{p_s^2}\left[\frac{(1-p_s^{(c)})p_s^{(c)}}{1-p_s^{(c)D}}p^{(c)}p_s^{(1)}p^{(1)}\right], \hfill \mbox{if~} c\neq 1
    \end{cases},
  \end{aligned}
  \nonumber
\end{equation}
where $\mu$ represents the corresponding expectations of all the variables in $\{\bar{k}_s^{(c)}, \bar{\chi}_s^{(c)}\}_{c\in C}$.
Denote
\begin{equation}
  \begin{aligned}
    A^{(c)}&=\frac{1}{p_s^2}\frac{(1-p_s^{(1)})^2}{1-p_s^{(1)D}}p^{(1)}p_s,\\
    B^{(c)}&=\frac{1}{p_s^2}\frac{(1-p_s^{(c)})^2}{1-p_s^{(c)D}}p^{(c)}p_s^{(1)}p^{(1)},\\
    C^{(c)}&=\frac{1}{p_s^2}\frac{(1-p_s^{(1)})p_s^{(1)}}{1-p_s^{(1)D}}p^{(1)}p_s,\\
    D^{(c)}&=\frac{1}{p_s^2}\frac{(1-p_s^{(c)})p_s^{(c)}}{1-p_s^{(c)D}}p^{(c)}p_s^{(1)}p^{(1)}.
  \end{aligned}
\end{equation}

Then 
\begin{equation}
  \begin{aligned}
    h_{\bar{k}_s^{(c)}}'&=
    \begin{cases}
      A^{(c)}-B^{(c)}, \hfil \mbox{if~} c=1\\
      -B^{(c)}, \hfill \mbox{if~} c\neq 1\\
    \end{cases},\\
    h_{\bar{\chi}_s^{(c)}}'&=
    \begin{cases}
      -C^{(c)}+D^{(c)}, \hfil \mbox{if~} c=1\\
      D^{(c)}, \hfill \mbox{if~} c\neq 1
    \end{cases}.
  \end{aligned}
\end{equation}

Since the chains for different classes are independent, for all $c\neq c'$ we have 
\[{\rm Cov}[\bar{k}_s^{(c)}, \bar{k}_s^{(c')}]={\rm Cov}[\bar{\chi}_s^{(c)}, \bar{\chi}_s^{(c')}]={\rm Cov}[\bar{k}_s^{(c)}, \bar{\chi}_s^{(c')}]=0.\]

Define
\begin{equation}
  \begin{aligned}
    \tau^2&=\sum_{c\in C}\sum_{c'\in C}\left[h_{\bar{k}_s^{(c)}}'h_{\bar{k}_s^{(c')}}'{\rm Cov}[\bar{k}_s^{(c)},\bar{k}_s^{(c')}]+h_{\bar{\chi}_s^{(c)}}'h_{\bar{\chi}_s^{(c')}}'{\rm Cov}[\bar{\chi}_s^{(c)},\bar{\chi}_s^{(c')}]\right.\\
      &\quad \left.+2h_{\bar{k}_s^{(c)}}'h_{\bar{\chi}_s^{(c')}}'{\rm Cov}[\bar{k}_s^{(c)},\bar{\chi}_s^{(c')}]\right]\\
      &=\sum_{c\in C}\left[h_{\bar{k}_s^{(c)}}'^2{\rm Var}[\bar{k}_s^{(c)}]+h_{\bar{\chi}_s^{(c)}}'^2{\rm Var}[\bar{\chi}_s^{(c)}]+2h_{\bar{k}_s^{(c)}}'h_{\bar{\chi}_s^{(c)}}'{\rm Cov}[\bar{k}_s^{(c)}, \bar{\chi}_s^{(c)}]\right]\\
      &=\sum_{c\in C}\left[B^{(c)2}{\rm Var}[\bar{k}_s^{(c)}]+D^{(c)2}{\rm Var}[\bar{\chi}_s^{(c)}]-2B^{(c)}D^{(c)}{\rm Cov}[\bar{k}_s^{(c)}, \bar{\chi}_s^{(c)}]\right]\\
      & \quad +\left[A(A-2B^{(1)}){\rm Var}[\bar{k}_s^{(1)}]+C(C-2D^{(1)}){\rm Var}[\bar{\chi}_s^{(c)}]\right.\\
      & \quad \left.+2(AD^{(1)}+B^{(1)}C-AC){\rm Cov}[\bar{k}_s^{(1)}, \bar{\chi}_s^{(1)}]\right]\\
      &=\left[\frac{p_s^{(1)}p^{(1)}}{p_s^2}\right]^2\sum_{c\in C}\frac{[1-p_s^{(c)}]^2p_s^{(c)}}{1-p_s^{(c)D}}p^{(c)2}
      +\left[\frac{p^{(1)}}{p_s^2}\right]^2 \frac{[1-p_s^{(1)}]^2p_s^{(1)}}{1-p_s^{(1)D}}p_s(p_s-2p_s^{(1)}p^{(1)}).
  \end{aligned}
\end{equation}

According to the Multivariate Delta Method, noticing that $\left.h\right|_\mu=q_s$, we can conclude that 
\[\sqrt{M}[\hat{q}_s^{(1)}-q_s]=\sqrt{M}[h(\{\bar{k}_s^{(c)}, \bar{\chi}_s^{(c)}\}_{c\in C})-q_s]\to n(0, \tau^2) \]
in distribution.
\hfill\BlackBox

\vskip 0.2in
\bibliography{ref}

\end{document}